\documentclass[10pt,twocolumn,letterpaper]{article}

 \usepackage[pagenumbers]{cvpr} 

\definecolor{cvprblue}{rgb}{0.21,0.49,0.74}
\usepackage[pagebackref,breaklinks,colorlinks,allcolors=cvprblue]{hyperref}

\def\paperID{8096} 
\def\confName{CVPR}
\def\confYear{2025}

\title{DECOR: Decomposition and Projection of Text Embeddings \\for Text-to-Image Customization}

\author{
    Geonhui Jang\textsuperscript{1}\footnotemark[1] , 
    Jin-Hwa Kim\textsuperscript{2,5}, 
    Yong-Hyun Park\textsuperscript{4}, 
    Junho Kim\textsuperscript{2}, 
    Gayoung Lee\textsuperscript{2}, 
    Yonghyun Jeong\textsuperscript{3}\footnotemark[2]\\[0.5em]
    \textsuperscript{1}School of Industrial and Management Engineering, Korea University, \textsuperscript{2}NAVER AI Lab, \\ \textsuperscript{3}NAVER Cloud, \textsuperscript{4}Department of Physics Education, Seoul National University, \textsuperscript{5}SNU AIIS\\[0.5em]
    \parbox{\textwidth}{
        \centering
        {\tt\small csleivear1@korea.ac.kr, enkeejunior1@snu.ac.kr, \\\{j1nhwa.kim, jhkim.ai, gayoung.lee, yonghyun.jeong\}@navercorp.com}
    }
}

\renewcommand{\thefootnote}{\fnsymbol{footnote}}

\begin{document}
\twocolumn[{%
\renewcommand\twocolumn[1][]{#1}%
\maketitle
\vspace{-30pt}
\begin{center}
    \centering
    \captionsetup{type=figure}
    \includegraphics[width=0.733\linewidth]{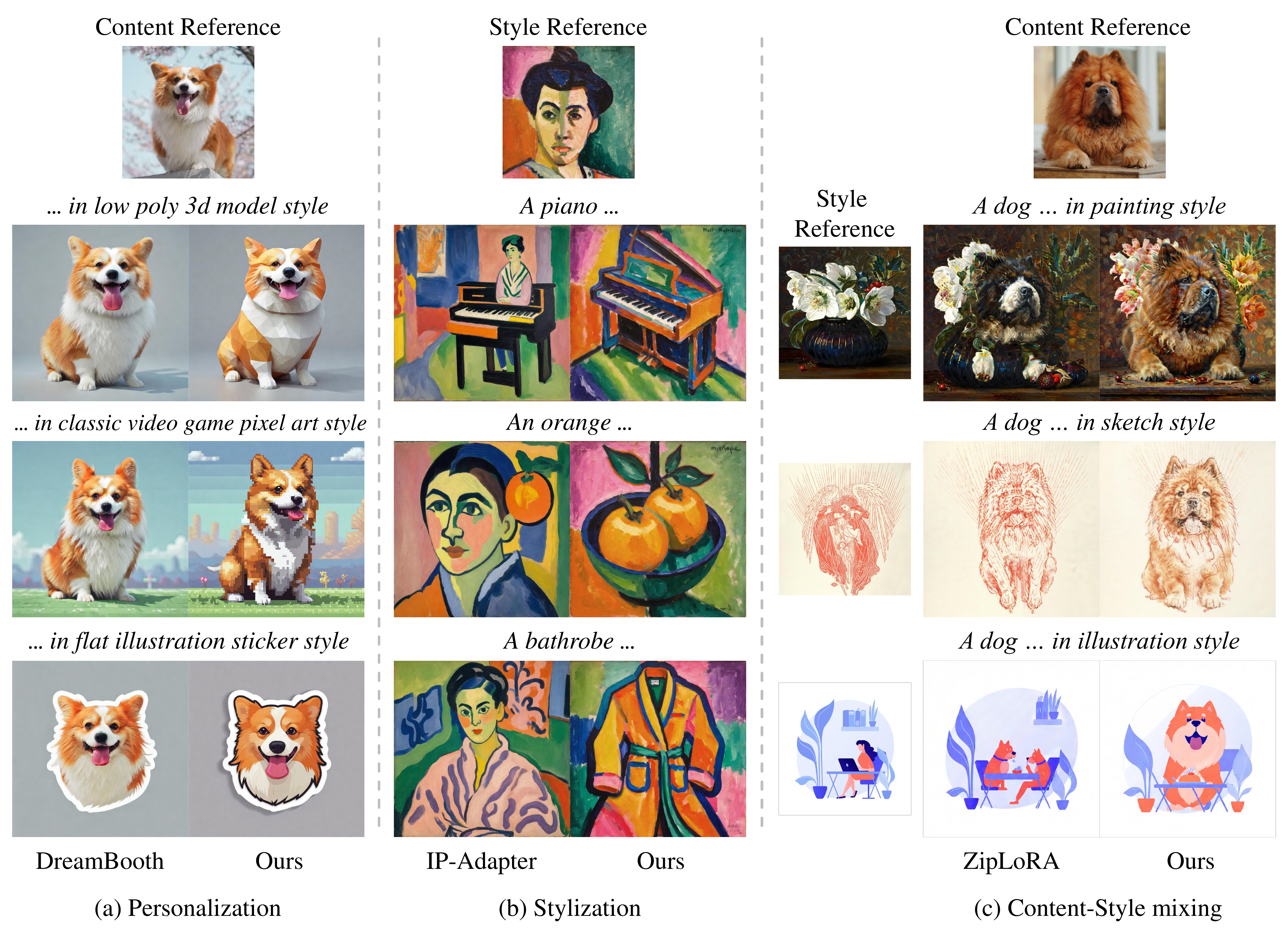}
    \vspace{-10pt}
    \caption{Our DECOR improves generation quality across personalization, stylization, and content-style mixing customization tasks.}
    \label{fig:fig1_teaser_image}
\end{center}
}]
\footnotetext[1]{First Author. Work done during an internship at NAVER Cloud.}
\footnotetext[2]{Corresponding Author.}
\renewcommand{\thefootnote}{\arabic{footnote}}

\begin{abstract}
Text-to-image (T2I) models can effectively capture the content or style of reference images to perform high-quality customization. A representative technique for this is fine-tuning using low-rank adaptations (LoRA), which enables efficient model customization with reference images. However, fine-tuning with a limited number of reference images often leads to overfitting, resulting in issues such as prompt misalignment or content leakage. These issues prevent the model from accurately following the input prompt or generating undesired objects during inference.
To address this problem, we examine the text embeddings that guide the diffusion model during inference. This study decomposes the text embedding matrix and conducts a component analysis to understand the embedding space geometry and identify the cause of overfitting. Based on this, we propose \textsc{DECOR}, which projects text embeddings onto a vector space orthogonal to undesired token vectors, thereby reducing the influence of unwanted semantics in the text embeddings. Experimental results demonstrate that \textsc{DECOR} outperforms state-of-the-art customization models and achieves Pareto frontier performance across text and visual alignment evaluation metrics. Furthermore, it generates images more faithful to the input prompts, showcasing its effectiveness in addressing overfitting and enhancing text-to-image customization.
\end{abstract}    
\section{Introduction}
\label{sec:intro}

Text-to-image (T2I) generation models are widely used in various fields of image generation. They can perform customization tasks such as personalization~\cite{dreambooth, textualinversion}, stylization~\cite{styledrop, ipadapter}, and content-style mixing~\cite{ziplora}, as shown in \cref{fig:fig1_teaser_image}. Personalization combines reference objects with unseen descriptions to generate images, stylization transfers the style of a reference image to new objects, and content-style mixing merges both tasks to depict a specific object in a specific style. These tasks typically use between one and five few-shot reference images.

Typically, customization is achieved through low-rank adaptation (LoRA), a parameter-efficient fine-tuning (PEFT) method that does not require retraining the entire model~\cite{lora, peft}. LoRA works by freezing the existing weight matrices and training only two low-rank matrices for each weight, improving the efficiency of the tuning process.

While LoRA tuning effectively updates the model, training T2I models with only a few reference images can lead to overfitting issues such as prompt misalignment and content leakage. 
\cref{fig:fig1_teaser_image} illustrates the issues of prompt misalignment and content leakage that occur when fine-tuning T2I models using reference images. As shown in (a) DreamBooth~\cite{dreambooth}, prompt misalignment can be observed where the model fails to follow the given prompt. In (b) IP-Adapter~\cite{ipadapter}, content leakage is observed, where undesired elements from the reference image appear in the generated output. Similarly, (c) ZIPLoRA~\cite{ziplora} also exhibits issues of prompt misalignment and content leakage.



We observed through extensive experiments that the issues of prompt misalignment and content leakage in T2I customization tasks are primarily rooted in the text condition, which guides the sampling process. By hierarchically decomposing the text embedding matrix, we identified that overfitting predominantly arises from the entanglement of word tokens with reference images. To address this challenge, we propose a novel framework called DECOmposition and pRojection (DECOR).

DECOR focuses on mitigating overfitting of word tokens by employing a projection-based refinement in the text embedding space. Specifically, our approach suppresses the influence of undesired token embeddings by projecting text embeddings in a direction orthogonal to these tokens. Experimentally, we demonstrate that this orthogonal projection effectively reduces their impact on the diffusion model’s output, thereby alleviating prompt misalignment and content leakage.

Notably, DECOR achieves these improvements without requiring additional training, making it computationally efficient. To the best of our knowledge, DECOR is the first method to conduct a detailed analysis of the text embedding space in the context of T2I customization. Comprehensive evaluations validate that our framework significantly outperforms existing state-of-the-art methods, effectively addressing prompt misalignment and content leakage while enhancing overall performance in customization tasks.

The significant contributions of this research are as follows\footnote{The source code will be available upon publication.}:

\begin{itemize}
    \item We analyze the causes of overfitting in T2I customization tasks, highlighting word tokens as the primary cause of prompt misalignment and content leakage.
    \item We propose a projection-based embedding refinement framework that mitigates the influence of undesired tokens on text embeddings without requiring additional training.
    \item Through extensive evaluations, we demonstrate that DECOR effectively addresses overfitting issues and achieves state-of-the-art performance in T2I customization tasks.
\end{itemize}
\section{Related work}
\label{sec:related_works}

\subsection{Customization with T2I models} 
The advent of Text-to-Image (T2I) diffusion models~\cite{imagen,stable_diffusion,sdxl,dalle} has revolutionized image generation, enabling unprecedented scalability and customization capabilities. These models excel at generating personalized objects and styles, significantly driven by advancements in Parameter Efficient Fine-Tuning (PEFT)~\cite{peft1, peft2, peft3}. Initial methods, such as Textual Inversion~\cite{textualinversion} and DreamBooth~\cite{dreambooth}, laid the groundwork by focusing on learning customized representations from user-provided data. Building on this foundation, approaches like Custom Diffusion~\cite{customdiffusion} introduced mechanisms to simultaneously learn multiple concepts, while SVDiff~\cite{svdiff} leveraged matrix decomposition to optimize learning within a compact parameter space.

Stylization in T2I models has also progressed significantly. Methods such as StyleDrop~\cite{styledrop} integrated adapters with Muse~\cite{muse} to facilitate customization, albeit requiring human feedback. Alternatively, training-free stylization techniques, such as StyleAligned~\cite{stylealigned} and Visual Style Prompting~\cite{vsp}, manipulate self-attention to maintain consistent styles across images without additional training. Recent innovations, like ZipLoRA~\cite{ziplora} and Break-for-Make~\cite{breakformake}, have proposed methods for merging LoRA weights for content and style customization.
Meanwhile, ours can apply to stylization, personalization, and their combination, enabling seamless integration into existing systems without the need for additional training.

\subsection{Mitigating overfitting for customization}

Overfitting remains a critical challenge in T2I customization, particularly with limited training data. Existing methods have proposed various solutions. FastComposer~\cite{fastcomposer} mitigates overfitting by employing delayed subject conditioning, while the Mixture-of-Attention (MoA)~\cite{moa} approach balances base and personalization attention to preserve prior knowledge. Perfusion~\cite{perfusion}, inspired by~\cite{locatinggpt}, constrains personalized subjects to adhere to their broader categorical context, reducing overfitting. Similarly, Infusion~\cite{infusion} considers the distribution of pre-trained models during training to address this issue.

Our approach offers a novel perspective by directly addressing overfitting through training-free modifications of text embeddings. By identifying spaces in word token embeddings that cause overfitting and projecting them onto orthogonal vectors, we eliminate content leakage and distorted image synthesis during customization. 
\section{Method}
\label{sec:method}

\setlength{\abovedisplayskip}{5pt} 
\setlength{\belowdisplayskip}{5pt} 

\setlength{\belowcaptionskip}{-15pt}

\subsection{Analysis on the CLIP text embedding space}
\label{sub_sec:3_1}

\begin{figure}[t]
    \centering
    \includegraphics[width=1.0\linewidth]{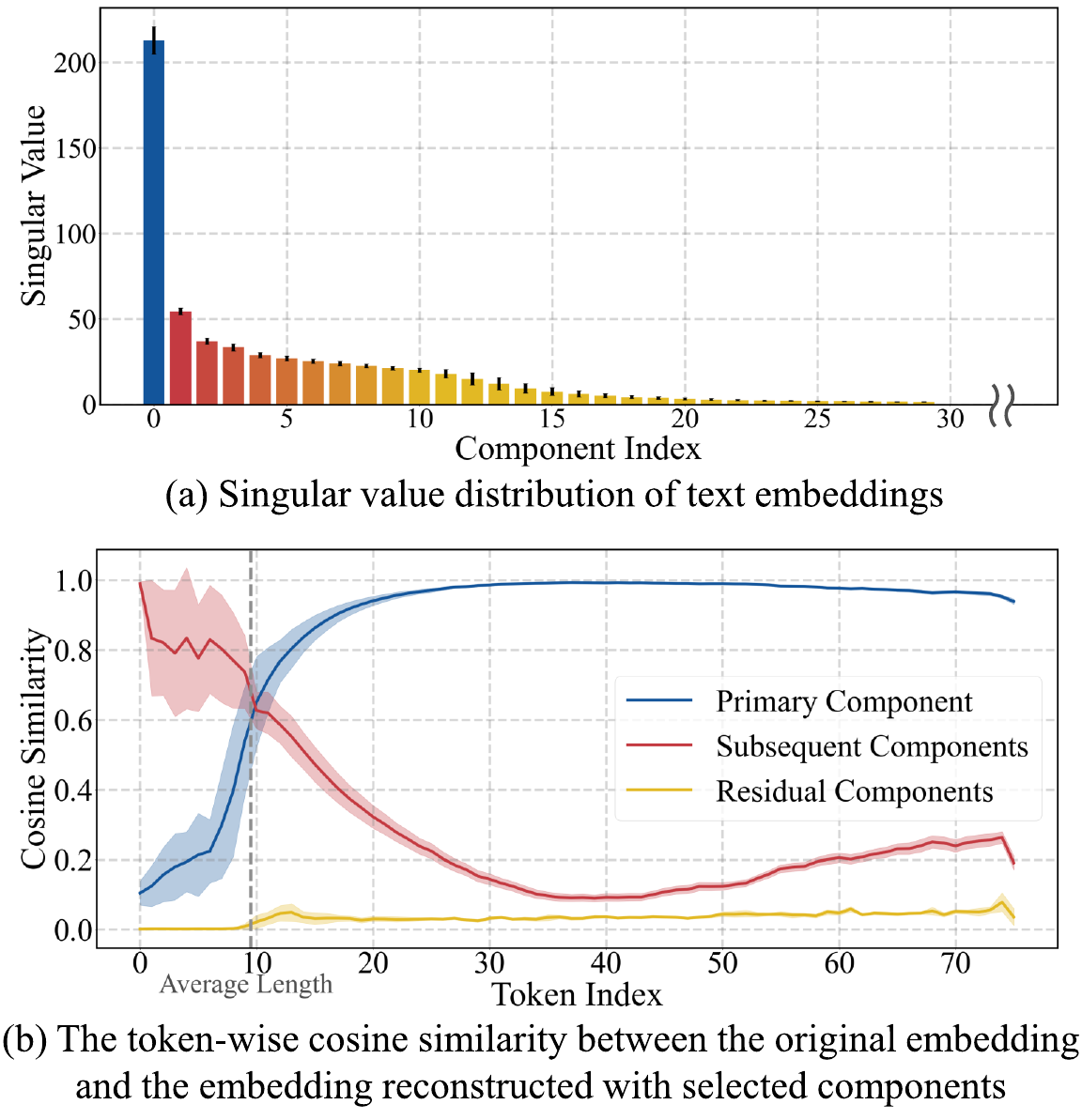}
    \caption{(a) The CLIP text embeddings have a large first singular value due to the high similarity of the [PAD] tokens. (b) The pattern of embedding reconstruction differs according to the magnitude of the singular values.}
    \label{fig:fig2_sv_and_emb_analysis}
\end{figure}

\noindent \textbf{Preliminary.} The CLIP text encoder~\cite{clip} converts an input prompt into a text embedding $X$, which can be defined as $X=\{X_\textnormal{w}; X_\textnormal{[PAD]}\} \in \mathbb{R}^{l \times d}$, where $l$ is the maximum length and $d$ is the embedding dimension. Here, $X_\textnormal{w} \in \mathbb{R}^{n \times d}$ and $X_\textnormal{[PAD]} \in \mathbb{R}^{(l-n) \times d}$ represent the embedding of the $n$ word tokens and the $(l-n)$ padding ([PAD]) tokens, respectively. Generally, CLIP text embeddings are set to $l=77$. The [PAD] tokens are added after the word tokens to fulfill the maximum length. We include padding tokens for this analysis because they are involved in the transformer attention mechanism. In this work, the `start of text' token is omitted and not considered.

\begin{figure}[t]
    \centering
    \hspace{-0.02\linewidth}
    \includegraphics[width=1.01\linewidth]{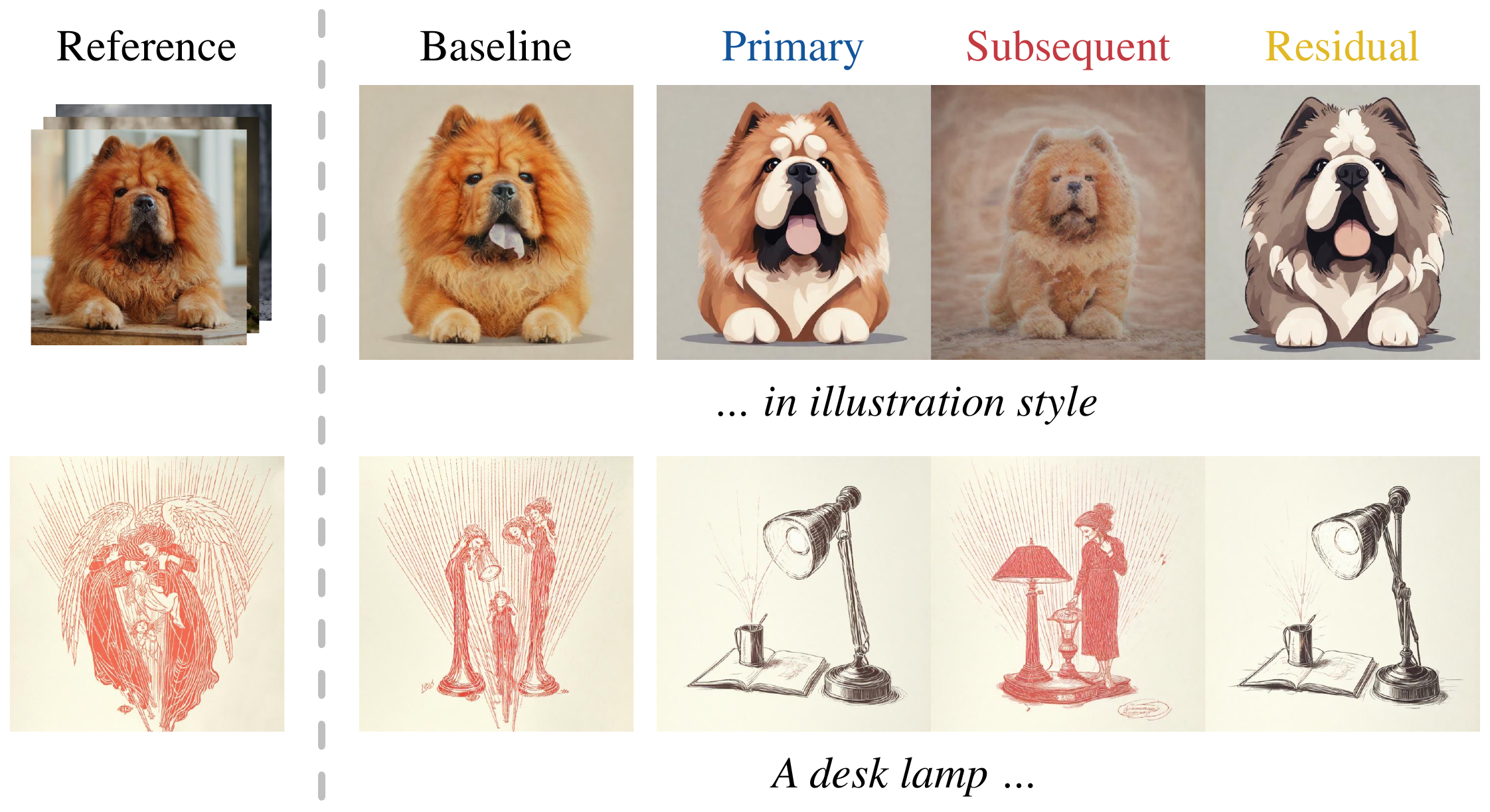}
    \caption{Customization results with the original embeddings (baseline) and the embeddings reconstructed using selected components (others).}
    \label{fig:fig3_components_compare}
\end{figure}

\noindent \textbf{Hierarchical structure of text embeddings.} First, we analyze the structure of the CLIP text embedding space. Understanding the geometry of embeddings is important for uncovering the intrinsic structure of the data. Since singular value decomposition (SVD)~\cite{svd} effectively isolates key components in the embedding space, we use it to decompose hierarchical embedding features along geometric axes. We apply SVD to $X$ as follows:
\begin{equation}
  X = U \Sigma V^T = \sigma_1 u_1 v_1^T + \sigma_2 u_2 v_2^T + \ldots + \sigma_l u_l v_l^T,
  \label{eq:1_svd}
\end{equation}
where $U \in \mathbb{R}^{l \times l}$ and $V \in \mathbb{R}^{d \times l}$ are orthonormal matrices representing the singular vectors, $\Sigma \in \mathbb{R}^{l \times l}$ is a diagonal matrix containing the singular values where $\sigma_1 > \ldots > \sigma_l$. The two graphs in \cref{fig:fig2_sv_and_emb_analysis} show the results of applying SVD to analyze the text embeddings for 20 prompts of a similar length; examples of these prompts can be found in the appendix. \cref{fig:fig2_sv_and_emb_analysis} (a) shows the average distribution of the first 30 singular values, $\{\sigma_i\}_{i=1}^{30}$, of the text embeddings. We can find that the first singular value, $\sigma_1$, is relatively large. This is because the [PAD] tokens that make up a significant portion of the text tokens tend to point in similar directions~\cite{getwhatyouwant}, leading the first singular vector to capture this common direction. \cref{fig:fig2_sv_and_emb_analysis} (b) shows the token-level cosine similarity between the original text embedding $X$ and the embeddings reconstructed by three specific component groups. We define the primary component as a single component embedding with the largest singular value, $\sigma_1 u_1 v_1^T$; the subsequent components as the cumulative embedding of the 3-10\% components ($i=2$ to $9$); and the residual components as the cumulative embedding of the 25-70\% components ($i=20$ to $54$). Specifically, the cosine similarity is calculated between each $d$-dimensional token vector in $X$ and the $d$-dimensional token vector of the embedding reconstructed by each component, across the $l$ total indices. The primary component (blue) accurately reconstructs $X_\textnormal{[PAD]}$ as mentioned above. The subsequent components (red) capture the information of the word token embeddings $X_\textnormal{w}$ that the first component does not explain. Lastly, the residual components (yellow) reconstruct noise in the embedding, with most token indices showing low similarity to the original embedding $X$.

\noindent \textbf{Customization with amplified embeddings.} In addition to the above analysis, we examine how each component affects customized image generation by varying the input text embeddings. In \cref{fig:fig3_components_compare}, the Baseline column shows DreamBooth's overfitted results using the original text embedding, and the remaining three columns show results where each component embedding is fed into the LoRA layers. We scale up the embedding to match the size of the original text embedding, which also amplifies the effect of the modified embedding. First, the primary or residual components reduce overfitting to the reference but fail to adequately capture the identity or style of the subject because of information loss. On the other hand, the subsequent components cause strong overfitting and distortions. From this, we can find that the word token embeddings $X_\textnormal{w}$ that serve as signals for LoRA customization become overly entangled with the reference image, leading to overfitting in the generated images.

Based on the above analyses, we argue that for effective customization, it is essential to limit the influence of word tokens that may cause prompt misalignment or content leakage. In the next section, we introduce a projection-based text embedding adjustment method for this and propose a framework to modify cross-attention operations in LoRA-based customization.
 
\subsection{Embedding projection for semantic separation}
\label{sub_sec:3_2}

\begin{figure}[t]
    \centering
    \includegraphics[width=1.0\linewidth]{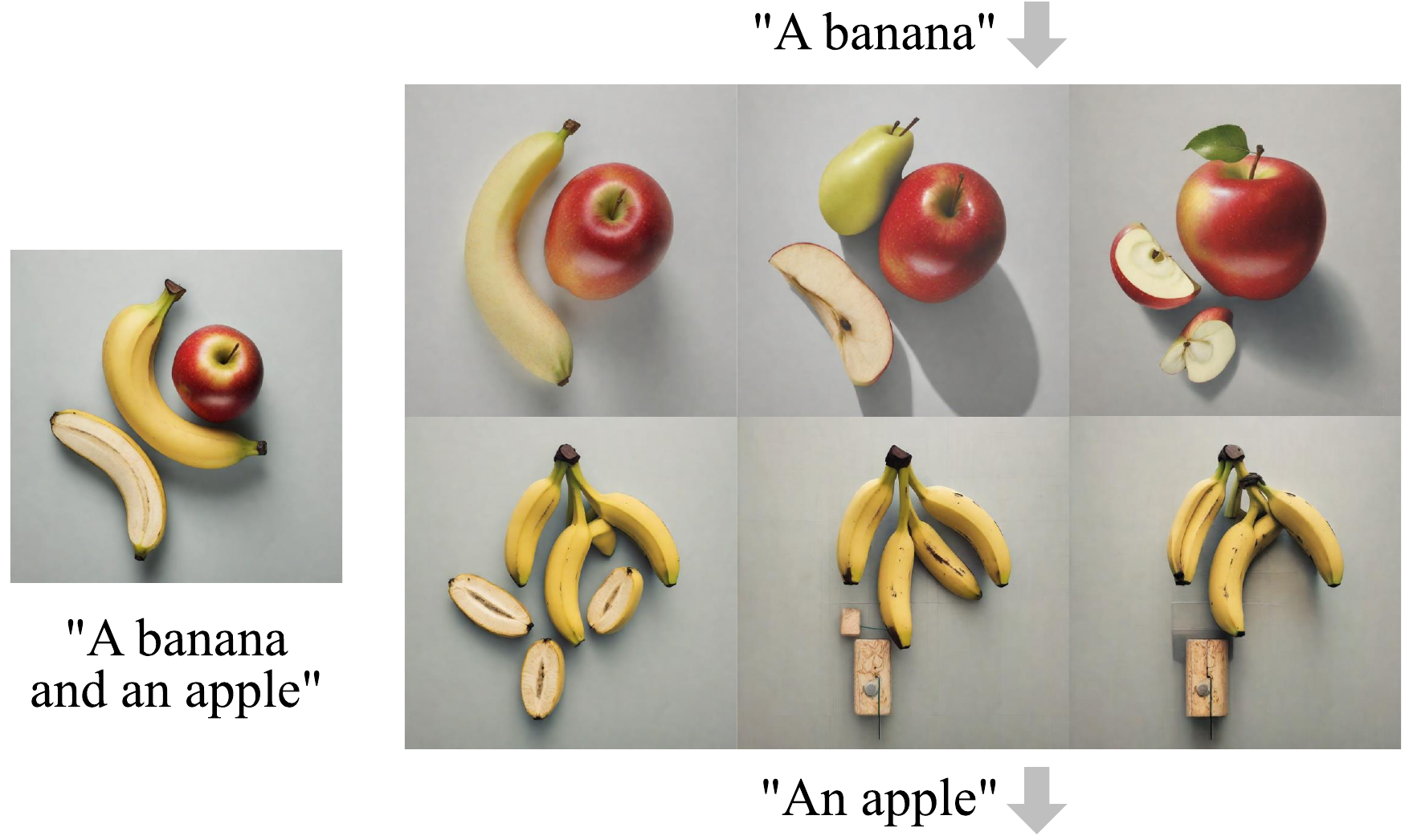}
    \caption{When the components along the axis of the unwanted word are subtracted from the original prompt embedding, this adjustment is reflected in the image generation results. From left to right, $\alpha$ is 0.5, 0.75, and 1.0.}
    \label{fig:fig4_apple_banana}
\end{figure}


\begin{figure}[t]
    \centering
    \hspace{-0.03\linewidth}
    \includegraphics[width=1.02\linewidth]{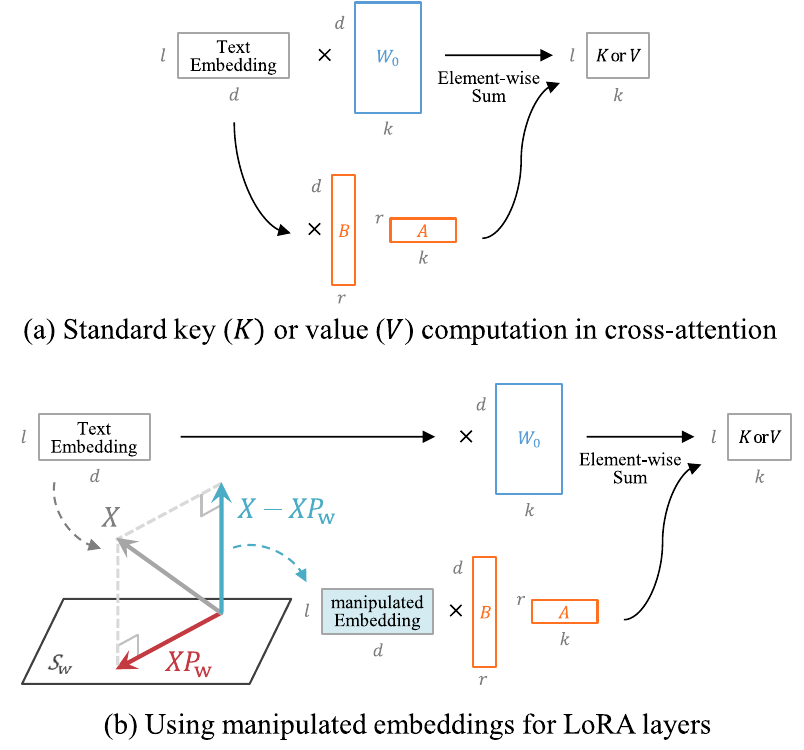}
    \caption{Comparison of the original and our inference pipeline. (a) In the standard approach, text embeddings are input into both the base and LoRA weights. (b) In the proposed method, we project the embedding onto word embedding space, and separate it from the original embedding. These manipulated embeddings are input into the LoRA layers, improving generation fidelity.}
    \label{fig:fig5_framework}
\end{figure}

\begin{figure}[b]
    \centering
    \hspace{-0.02\linewidth}
    \includegraphics[width=1.01\linewidth]{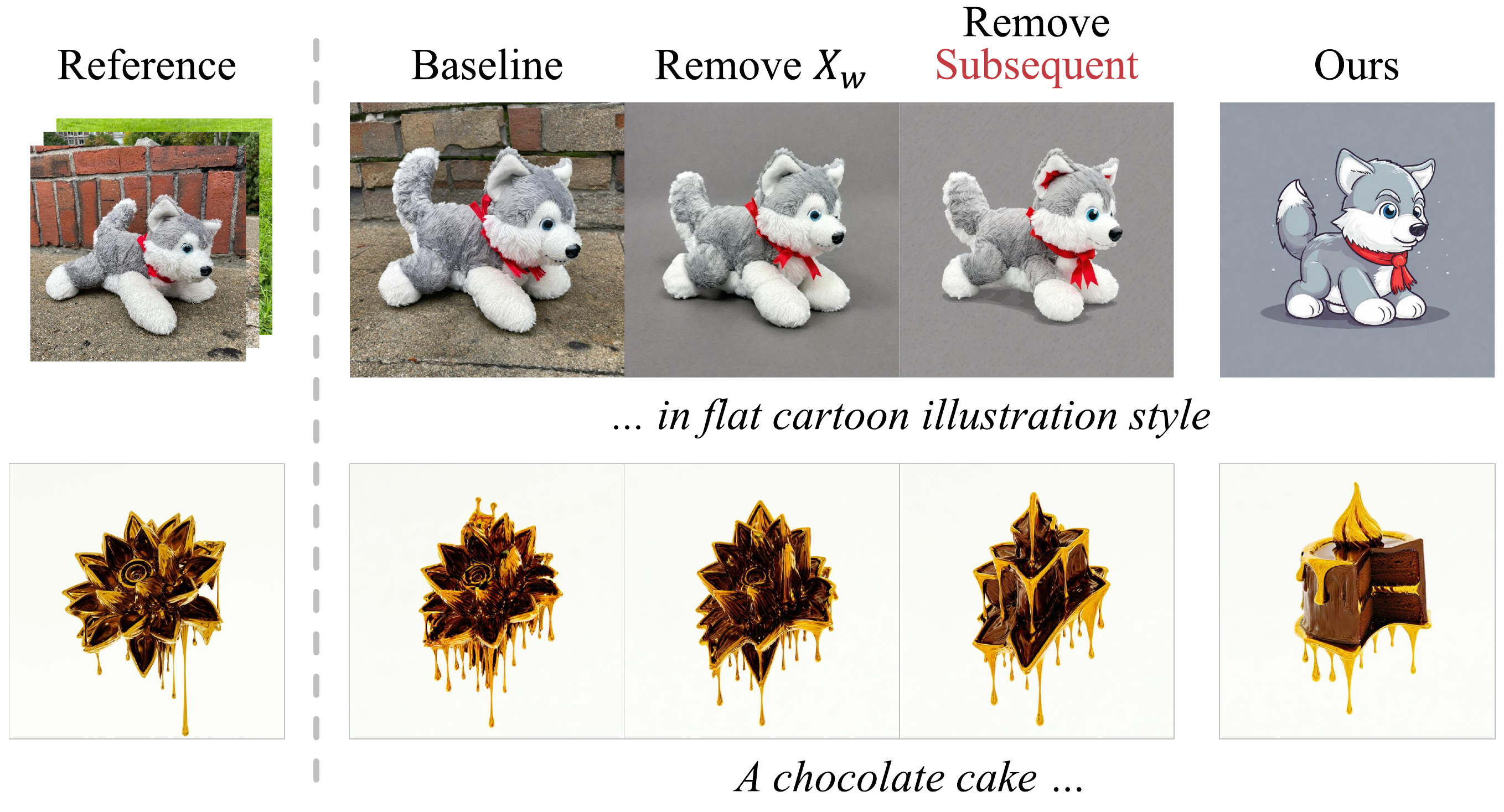}
    \caption{Simply removing the word embedding $X_\textnormal{w}$ from the original embedding or using an embedding reconstructed without the subsequent components cannot solve the overfitting problem.}
    \label{fig:fig6_suppression_comparison}
\end{figure}

In this section, we introduce a straightforward yet effective method to address the overfitting problem described in \cref{sub_sec:3_1}. We propose a projection technique that allows text embeddings to separate from undesired elements, based on the previous analysis of the CLIP text embedding space. Our method builds on existing studies~\cite{projection_human, projection_cls} that leverage the orthogonality of text embeddings in the latent space to project the text embedding matrix onto specific semantic axes. Let $\tilde{X}$ denote the embedding we aim to suppress, and $S_{\tilde{X}}$ as the subspace spanned by $\tilde{X}$. To separate the text embedding $X$ from $S_{\tilde{X}}$, we remove the component of $X$ that lies along the axis of $S_{\tilde{X}}$ as follows:

\begin{equation}
  X' = X - \alpha X P_{\tilde{X}},
  \label{eq:2_projection}
\end{equation}
where $P_{\tilde{X}} \in \mathbb{R}^{d \times d}$ is a projection matrix onto $S_{\tilde{X}}$, and $\alpha$ is a hyperparameter between 0 and 1.0 that controls the degree of removal of the projected feature. \cref{fig:fig4_apple_banana} shows the image generation results after modifying the text condition embedding using the projection. By defining $\tilde{X}$ as the embedding of each negative target and using the modified text embedding $X'$ as in \cref{eq:2_projection}, the target can be removed in the generated images. There are various methods, such as SVD or the Gram-Schmidt process~\cite{schmidt}, that can be used to obtain the projection matrix $P_{\tilde{X}}$. Here, we use SVD, where it is known that the projection matrix $P_{\tilde{X}}$ can be defined using the orthonormal matrix $\tilde{V}$, as follows:
\begin{equation}
  P_{\tilde{X}} = \tilde{V} \tilde{V}^T,
  \label{eq:3_P_tilde}
\end{equation}
where $\tilde{X} = \tilde{U} \tilde{\Sigma} \tilde{V}^T$. Through this projection and separation process, we can obtain an embedding in which the influence of $\tilde{X}$ is suppressed in $X$, which is consistent with the T2I generation results in \cref{fig:fig4_apple_banana}.

\noindent \textbf{Customization framework.} We propose a framework, DECOR, that can apply this projection technique to customization tasks. Typically, during training and inference using LoRA, the key and value computations in cross-attention are performed by feeding the text embeddings into the base and LoRA weights, as shown in \cref{fig:fig5_framework} (a). The DECOR framework follows the baseline process for training the LoRA layer, while inference is performed as illustrated in \cref{fig:fig5_framework} (b). The fine-tuned LoRA weight $\Delta W = BA$ receives the manipulated text embedding $X'=X - \alpha X P_{X_\textnormal{w}}$, which is computed through the aforementioned process. Before being input, $X'$ is resized to match the size of the original text embedding.

\noindent \textbf{Comparing suppression approaches.} As a simple way to reduce the influence of the word token embedding $X_\textnormal{w}$, one might consider suppressing $X_\textnormal{w}$ directly. \cref{fig:fig6_suppression_comparison} presents an experiment with changes related to $X_\textnormal{w}$. The first column shows the baseline results from DreamBooth, and the second column shows the results when the elements in the positions of $X_\textnormal{w}$ are set to zero. The third column presents the results after decomposing $X$ with SVD and reconstructing the embedding, excluding the subsequent components that capture $X_\textnormal{w}$ as described in \cref{sub_sec:3_1}. These naive approaches fail to address issues such as prompt misalignment and content leakage. In particular, since the subsequent components are identified by the order of singular vectors instead of token indices, it is unclear which components correspond to $X_\textnormal{w}$. In contrast, our approach, which uses projection onto semantic axes in the text embedding space, effectively adjusts the embedding and addresses overfitting issues.
\section{Experiments}
\label{sec:experiments}

\setlength{\belowcaptionskip}{-15pt}

DECOR is a versatile approach that can be applied to various LoRA-based customization tasks, such as personalization, stylization, and content-style mixing. To evaluate DECOR, we designed experiments for each of these three tasks and selected appropriate comparison methods.
\subsection{Experimental setup}
\noindent \textbf{Dataset.} For the personalization experiments, we used a subset of the DreamBooth dataset~\cite{dreambooth}, consisting of 12 subjects. For each subject, we trained the LoRA layers using 4–5 reference images. In the stylization experiments, we selected from the StyleDrop dataset~\cite{styledrop}, along with additional images that exhibit unique styles, for a total of 21 style reference images. For each style, we trained the LoRA layers using a single image. For the content-style mixing experiments, we combined 8 subjects from the personalization dataset with 12 styles from the stylization dataset, resulting in a total of 96 subject-style image pairs.

\noindent \textbf{Details.} Except for StyleDrop, which uses a ViT-based model~\cite{vit}, we used SDXL~\cite{sdxl}, a state-of-the-art T2I generative model. Since there is no official model checkpoint for StyleDrop, we used an open-source reproduction, the aMUSEd-512 model~\cite{amused}. All LoRA-based methods apply LoRA layers only to the main model, such as U-Net~\cite{unet} or ViT~\cite{vit}, and not to the text encoder. Unless otherwise specified, hyperparameter settings for all baseline methods follow the original papers. In all qualitative comparisons, we set $\alpha=0.8$, which provides the best results.

\begin{figure}[t]
  \centering
  \includegraphics[width=1.0\linewidth]{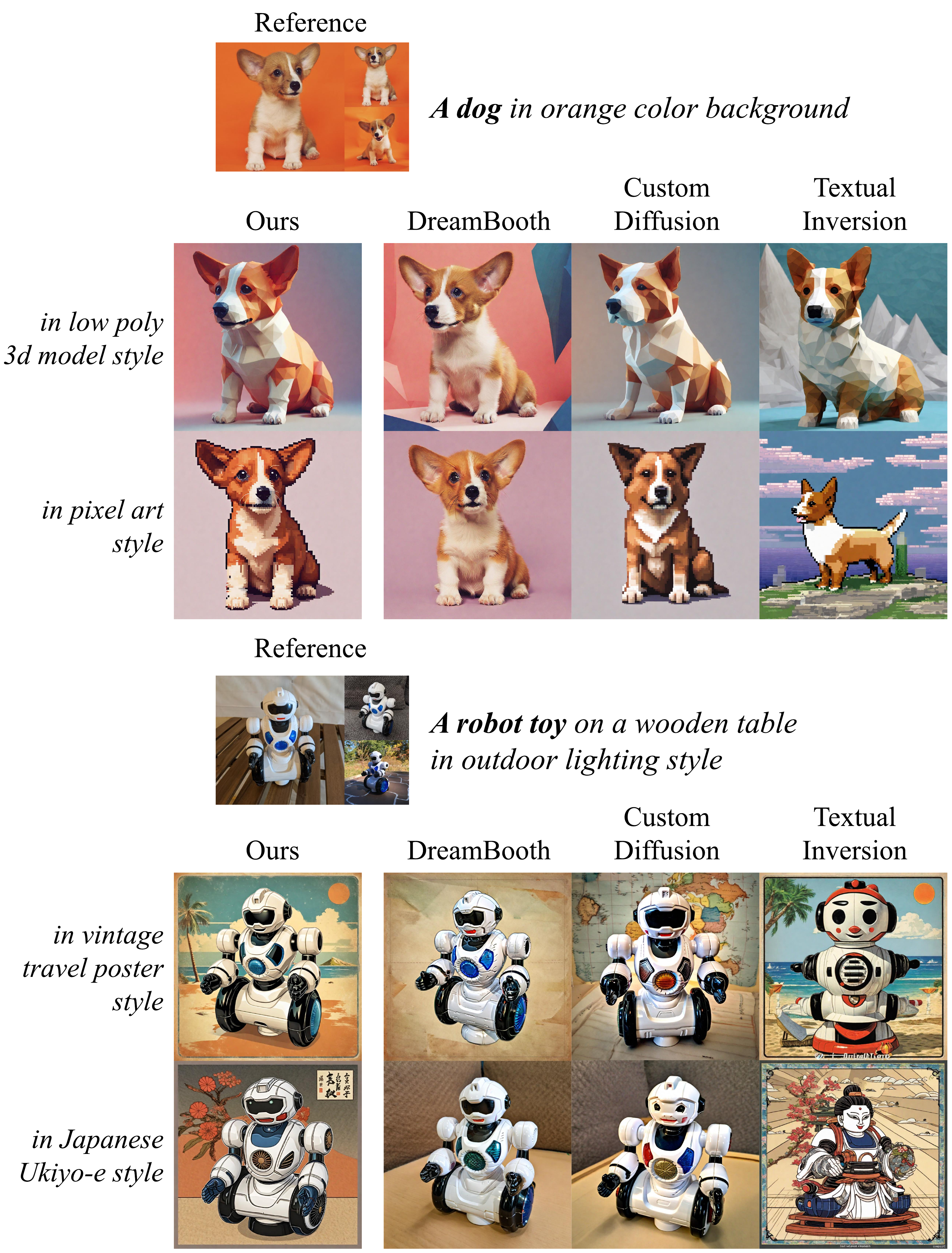}
  \caption{Qualitative personalization comparison.}
  \label{fig:fig7_personalization_results}
\end{figure}

\noindent \textbf{Evaluation metrics.} To assess the quality of generated images, we used CLIP image-text similarity (CLIP ViT-L/14)~\cite{clip} and DINO feature similarity (DINOv2 ViT-B/14)~\cite{dino, dinov2}, as previous studies~\cite{dreambooth, stylealigned, vsp}. These metrics are suitable for quantitatively evaluating customization performance for the following reasons: CLIP image-text similarity is defined as the cosine similarity between the CLIP image embedding of the generated image and the CLIP text embedding of the sampling prompt, making it appropriate for evaluating the text alignment. DINO feature similarity is defined as the average cosine similarity between the DINO feature embeddings of the reference image and the generated image. Because the DINO model is trained through self-supervised learning for image classification and feature extraction, this metric is well-suited for measuring identity preservation or style transfer performance.

\subsection{Personalization}
\label{sub_sec:4_1}

\noindent \textbf{Experimental setup.} For comparison methods, we selected DreamBooth, Textual Inversion, and Custom Diffusion~\cite{dreambooth, textualinversion, customdiffusion}. Because the proposed method focuses on improving text alignment, we defined artistic templates to evaluate how well the additional descriptions are reflected in the generated images. The artistic templates are sets of prompts designed to depict an object in specific artistic styles, such as ``{object} in cartoon illustration style'' or ``{object} in pixel art style.'' For each object, we used 20 artistic prompts for evaluation.

\noindent \textbf{Qualitative comparison.} \cref{fig:fig7_personalization_results} shows the results of personalized image generation. The prompts next to each reference image were used to train the LoRA layer, and new additional style descriptions were used to generate the images in each row. In personalization tasks, it is crucial to preserve the identity of the subject in the reference image while faithfully following the given prompt. In \cref{fig:fig7_personalization_results}, the proposed DECOR generates images that follow the additional unseen style descriptions. For example, for the dog at the top, DECOR effectively describes the 3D modeling or pixel art characteristics in the generated images. In contrast, DreamBooth overfits to the subject in the reference image, merely replicating the original one. Custom Diffusion and Textual Inversion exhibit lower image fidelity or fail to preserve the identity of the subject.

\begin{figure}[t]
    \centering
    \includegraphics[width=0.8\linewidth]{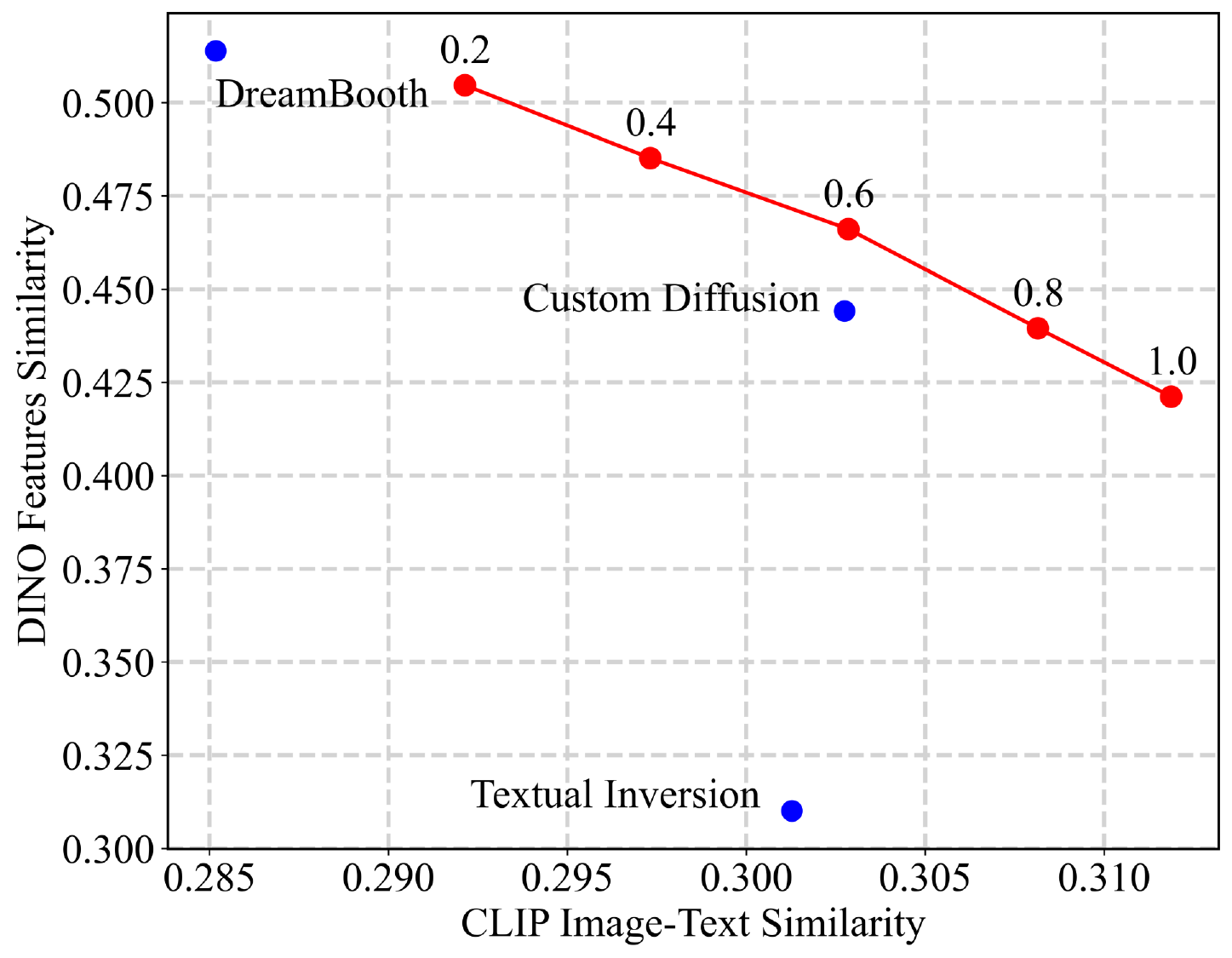}
    \caption{Quantitative personalization comparison. Our method demonstrates superior results regarding text fidelity and image features as $\alpha$ is varied.}
    \label{fig:fig8_personalization_score}
\end{figure}

\noindent \textbf{Quantitative comparison.} In \cref{fig:fig8_personalization_score}, DECOR demonstrates the Pareto optimality between text alignment (\ie, CLIP text-image similarity) and visual alignment (\ie, DINO features similarity) compared to other methods. By adjusting the hyperparameter $\alpha$, we can control the trade-off between these two metrics. DreamBooth shows the highest score in visual alignment, which is attributable to its tendency to merely replicate the reference subject, resulting in a very high similarity to the reference images, as shown in \cref{fig:fig7_personalization_results}.

\begin{figure}[t]
    \centering
    \vspace{1em}
    \includegraphics[width=1.0\linewidth]{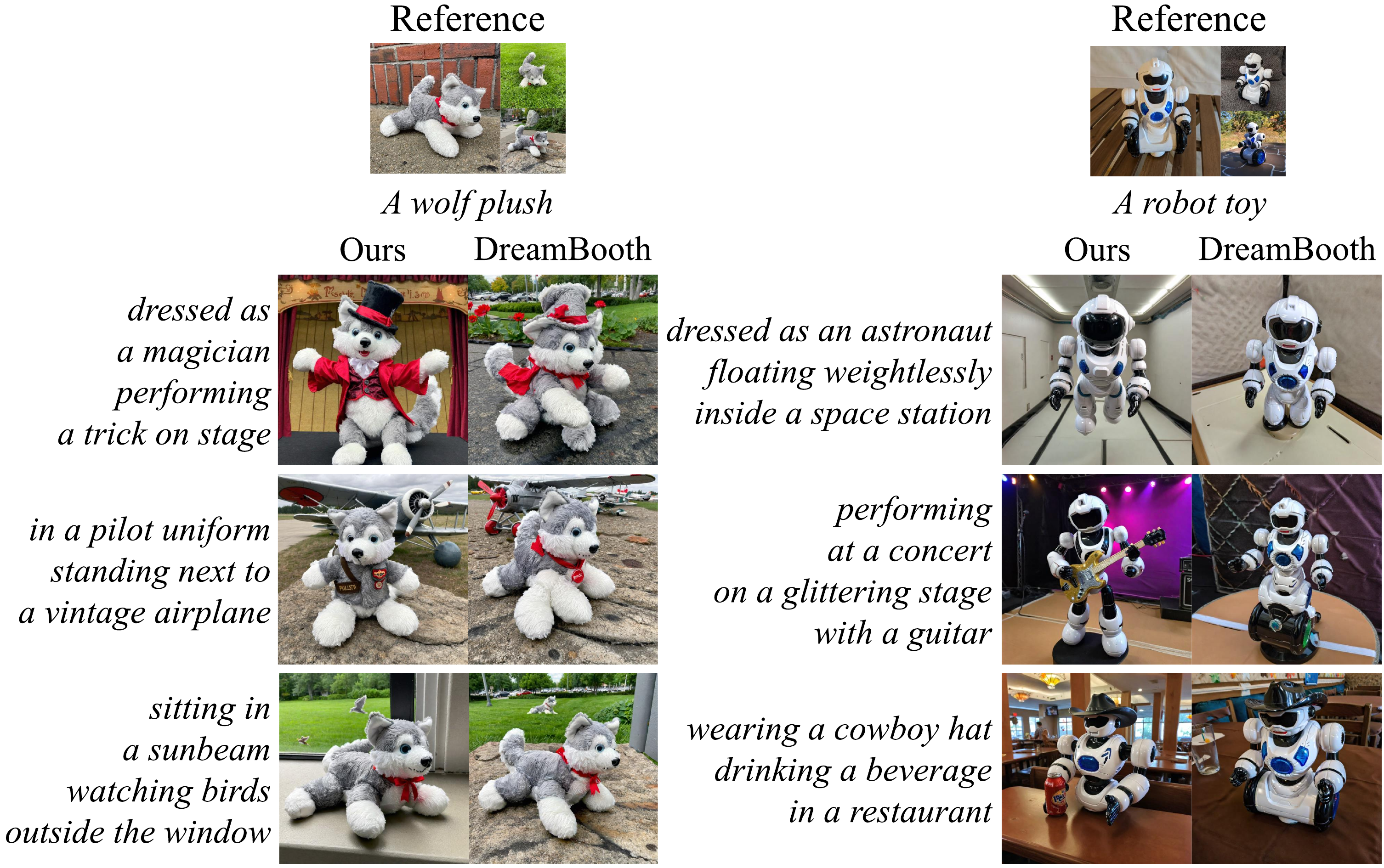}
    \caption{Personalization results using realistic templates.}
    \hspace{0.25cm}
    \label{fig:fig9_personalization_realistic_template}
\end{figure}

\noindent \textbf{Realistic template.} Although our primary focus is on text fidelity using artistic templates, DECOR also performs well with realistic templates that describe ordinary real-world scenarios. As shown in \cref{fig:fig9_personalization_realistic_template}, DECOR achieves comparable quality to the baseline DreamBooth on realistic templates, effectively preventing overfitting to the reference image and accurately following the given text while producing more diverse visual expressions. For example, as shown on the right side of \cref{fig:fig9_personalization_realistic_template}, DECOR successfully captures additional descriptions for ``A robot toy,'' reflecting details such as ``with a guitar'' in the second column. In contrast, the baseline method fails to capture these details accurately.

\subsection{Stylization}
\label{sub_sec:4_2}

\begin{figure}[t]
  \centering
  \includegraphics[width=1.0\linewidth]{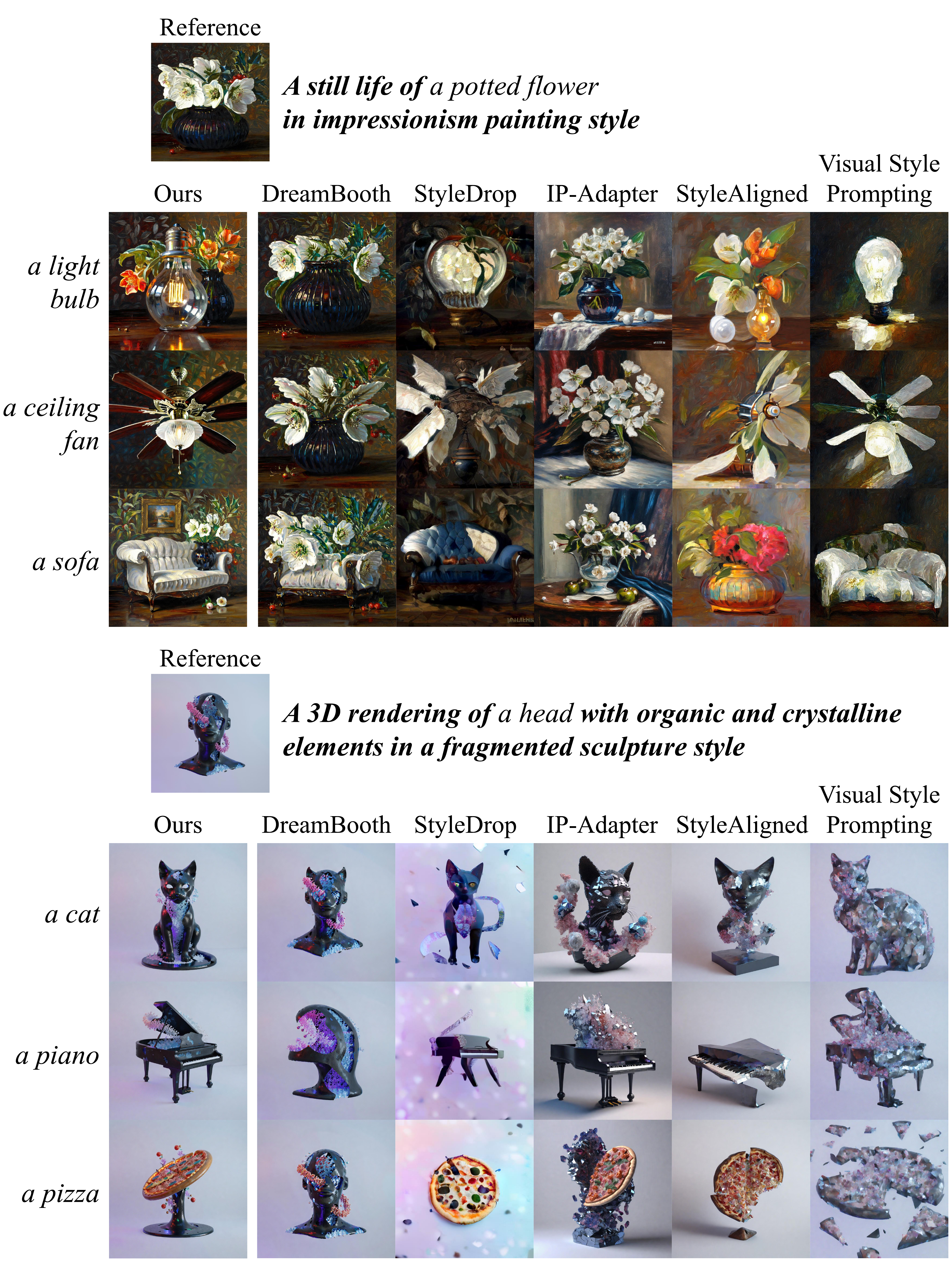}
  \caption{Qualitative stylization comparison.}
  \label{fig:fig10_stylization_results}
\end{figure}

\noindent \textbf{Experimental setup.} We selected DreamBooth, StyleAligned, Visual Style Prompting, IP-Adapter, and StyleDrop as comparison methods~\cite{dreambooth, stylealigned, vsp, ipadapter}. For each style, we evaluated the performance using 50 target objects. StyleAligned and Visual Style Prompting, which are designed for synthetic image style transfer, propose extensions for real images by obtaining latent features of real images using DDIM~\cite{ddim} and stochastic inversion~\cite{ddpm}, respectively. We followed these methods in our experiments. For a fair comparison, we used the same initial noise across all experiments except for StyleDrop because it is based on a ViT model. As noted in~\cite{offsetnoise}, adding a small random scalar to the Gaussian noise during training LoRA layers helps capture flat stylistic textures, which is suitable for creating illustrative styles. Therefore, we applied this noise offset technique, setting the offset scale to 0.1 for both DECOR and DreamBooth.

\noindent \textbf{Qualitative comparison.} \cref{fig:fig10_stylization_results} shows the results of the stylization comparison. First, we compared DECOR with training-based methods such as DreamBooth and StyleDrop. DECOR successfully transfers the style of the reference image while faithfully following the given prompts. In contrast, DreamBooth suffers from content leakage because of overfitting, and StyleDrop fails to accurately capture the style or exhibits poor text fidelity. By refining the text embeddings, DECOR prevents overfitting to the reference and improves text fidelity.

Next, we compared DECOR with training-free methods such as IP-Adapter, StyleAligned, and Visual Style Prompting. As shown in \cref{fig:fig10_stylization_results}, IP-Adapter has issues with content leakage from the reference image (\eg, the flowers), likely because of conflicts between the text and the image features condition in the IP-Adapter’s mechanism. StyleAligned and Visual Style Prompting, which use DDIM~\cite{ddim} and stochastic inversion~\cite{ddpm}, capture the overall color and text descriptions well, but generate somewhat distorted images. As noted in~\cite{nti, npi}, these inversion techniques often cause the image latent features to deviate from the optimal generation path, leading to a degradation of visual quality. In contrast, DECOR achieves a level of detailed synthesis that cannot be achieved by these training-free methods.

\begin{figure}[t]
    \centering
    \includegraphics[width=0.8\linewidth]{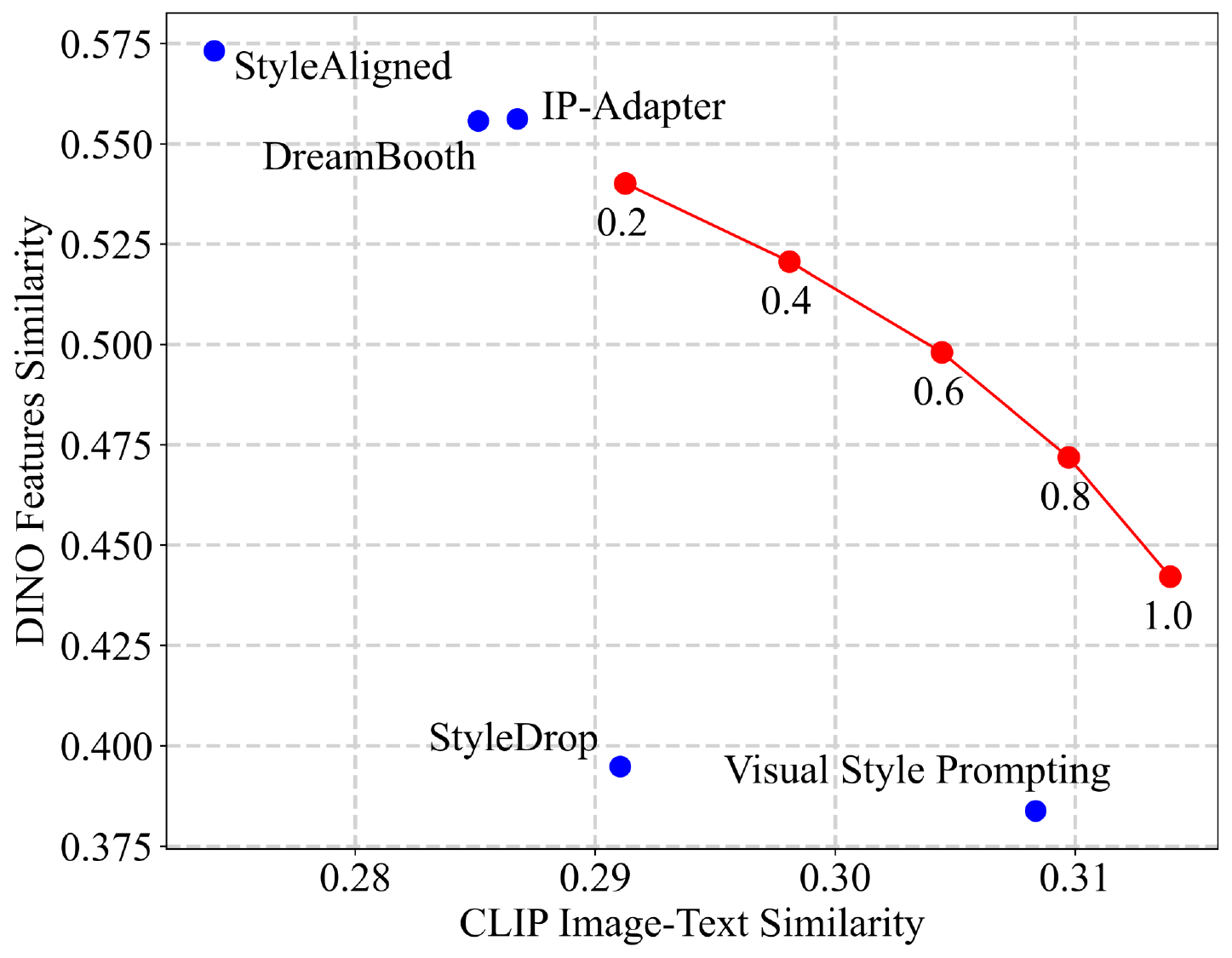}
    \caption{Quantitative stylization comparison. In contrast to other stylization methods that overly prioritize a single metric, our method exhibits Pareto optimality across both scores.}
    \label{fig:fig11_stylization_score}
\end{figure}

\noindent \textbf{Quantitative comparison.} \cref{fig:fig11_stylization_score} shows a quantitative comparison with the other methods. Similar to the personalization task, DECOR demonstrates an optimal trade-off between text alignment and visual alignment by adjusting the hyperparameter $\alpha$. As mentioned in~\cite{vsp}, StyleAligned exhibits poor performance in text alignment because of content leakage from the reference image.

\subsection{Content-Style mixing}
\label{sub_sec:4_3}

Typically, multiple fine-tuned LoRA layers can be merged simultaneously during inference in a simple additive manner because of their residual connection mechanism. However, as noted in~\cite{ziplora}, directly merging independently trained LoRA layers may cause weight conflicts, resulting in degraded quality. DECOR avoids this issue by refining semantics and reducing noise from the text embeddings.

\begin{figure}[t]
    \centering
    \includegraphics[width=1.0\linewidth]{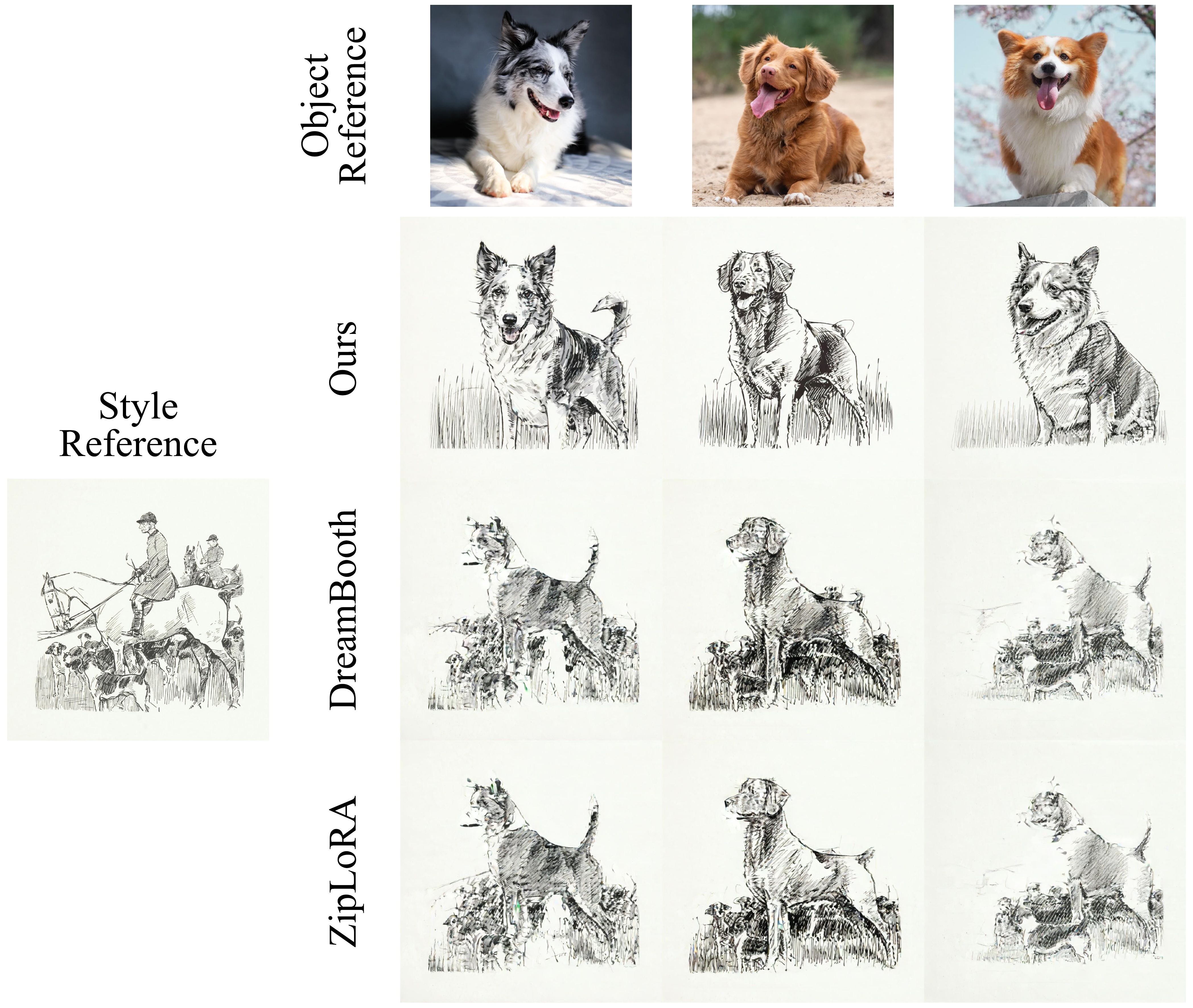}
    \caption{Qualitative content-style mixing comparison with LoRA merging methods.}
    \label{fig:fig12_mixing_results}
\end{figure}

\noindent \textbf{Experimental setup.} The content-style mixing task is based on merging content LoRA (i.e., personalization LoRA layers) and style LoRA (i.e., stylization LoRA layers). We compared DECOR with DreamBooth and ZipLoRA. In DECOR, we input different embeddings, such as the original embeddings or projected embeddings under different values of the hyperparameter $\alpha$, into the content and style LoRA to observe patterns. Because no official source code of ZipLoRA is available, we used unofficial implementation~\cite{ziprora_unofficial}. In all experiments, the LoRA layer scales are set to 1.0.

\noindent \textbf{Qualitative and Quantitative comparison.} As shown in \cref{fig:fig12_mixing_results}, DECOR expresses the target style without distortion. In contrast, DreamBooth, which directly merges content and style LoRA layers, fails to preserve the identity of the subject, and ZipLoRA does not effectively resolve conflicts between the two concepts. For quantitative results, please refer to the appendix.

\subsection{Ablation study and Visualization}
\label{sub_sec:4_4}

\begin{figure}[t]
    \centering
    \includegraphics[width=1.0\linewidth]{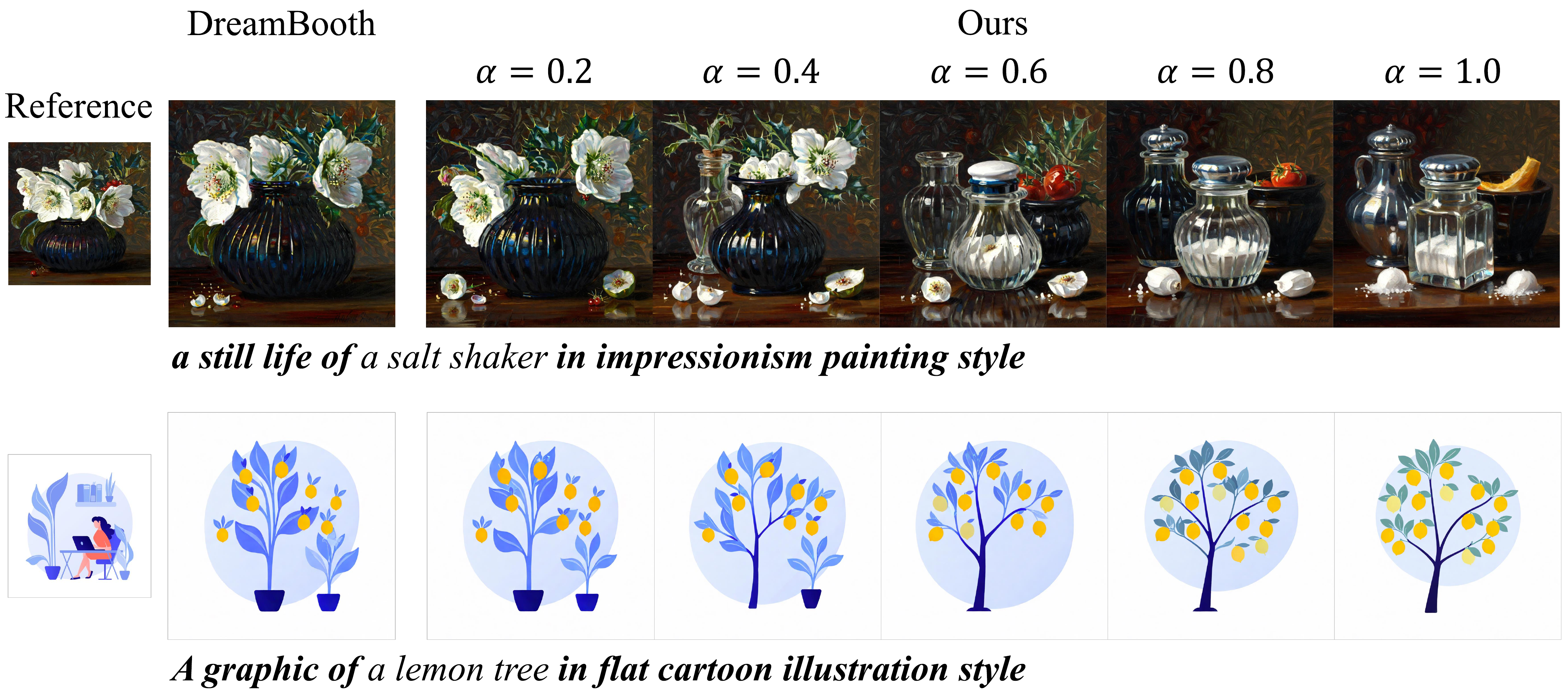}
    \caption{Ablation study on $\alpha$. In the stylization task, varying $\alpha$ reveals two key effects: preventing overfitting, such as content leakage (top), and controlling fine style components (bottom).}
    \label{fig:fig13_ablation}
\end{figure}

We discussed the method of using $\alpha$. By adjusting $\alpha$ in \cref{eq:2_projection}, we can control the degree to which unwanted token components are removed from the embedding. This adjustment offers two advantages in the stylization task: preventing content leakage from the reference image and controlling detailed components. The first row of \cref{fig:fig13_ablation} shows how controlling $\alpha$ helps prevent content leakage. The second row illustrates how adjusting $\alpha$ affects the intensity of finer stylistic details. By continuously adjusting $\alpha$, we obtained controllability over the stylization process, demonstrating DECOR's flexible applicability.

Interestingly, this controllability appears to vary depending on the complexity of the reference image. As shown in \cref{fig:fig13_ablation}, we found that for more complex visual expressions, such as oil painting style, DECOR tended to mitigate overfitting and content leakage, whereas for simpler, such as flat illustration style, it helps control detailed components. We attribute this to the fact that when the reference image is visually complex, the LoRA layer tends to become more strongly entangled with the text embeddings during training, leading to overfitting to the text condition and prone to content leakage.

\begin{figure}[t]
    \centering
    \vspace{1em}
    \includegraphics[width=1.0\linewidth]{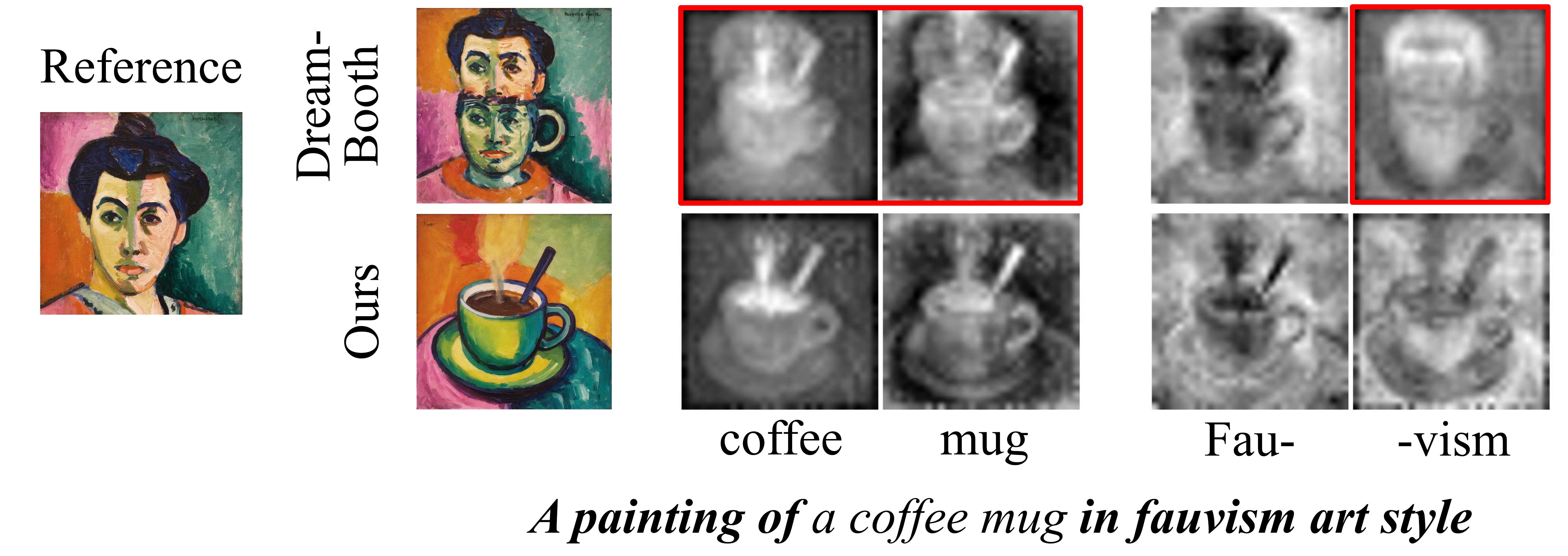}
    \caption{Visualization of attention map.}
    \label{fig:fig14_attention_map}
\end{figure}

\cref{fig:fig14_attention_map} visualizes the attention maps during the stylization process of the diffusion model. In DreamBooth, the attention map for the word ``fauvism'' shows that the token embeddings overly affect feature calculations during the attention operations, interfering with the generation of the target object, ``coffee mug.'' Additionally, the embedding for the word ``coffee mug'' fails to properly attend to the target pixels, causing misalignment. In contrast, the attention maps from our projection method demonstrate that both words effectively attend to the correct pixels without interfering with each other, allowing for precise image feature calculations.

\section{Conclusion}
\label{sec:conclusion}

We address the issues of prompt misalignment and content leakage in LoRA-based T2I customization tasks. We discover that these issues arise because the LoRA layers become too closely entangled with the text word embeddings, limiting the model's ability to accurately follow given prompts. To solve this, we introduce DECOR, an effective method that enhances text representations without additional training. DECOR uses projection on the text embeddings and separate  to emphasize the key semantics. This process highlights reduces unwanted feature in the text embeddings, leading to more faithful image generation. Our extensive evaluations demonstrate that DECOR outperforms state-of-the-art customization models, achieving optimal results in both visual similarity and text alignment scores. This study highlights the importance of understanding and the flexibility of adjusting the text embedding space in T2I models, especially when dealing with limited reference images. By providing a straightforward solution that does not require retraining, DECOR offers a practical improvement to the field of image generation. Future research could explore combining DECOR with other fine-tuning methods to enhance customization capabilities.

\clearpage

{
    \small
    \bibliographystyle{ieeenat_fullname}
    \bibliography{main}

\begin{thebibliography}{52}
\providecommand{\natexlab}[1]{#1}
\providecommand{\url}[1]{\texttt{#1}}
\expandafter\ifx\csname urlstyle\endcsname\relax
  \providecommand{\doi}[1]{doi: #1}\else
  \providecommand{\doi}{doi: \begingroup \urlstyle{rm}\Url}\fi

\bibitem[Achiam et~al.(2023)Achiam, Adler, Agarwal, Ahmad, Akkaya, Aleman, Almeida, Altenschmidt, Altman, Anadkat, et~al.]{gpt4}
Josh Achiam, Steven Adler, Sandhini Agarwal, Lama Ahmad, Ilge Akkaya, Florencia~Leoni Aleman, Diogo Almeida, Janko Altenschmidt, Sam Altman, Shyamal Anadkat, et~al.
\newblock Gpt-4 technical report.
\newblock \emph{arXiv preprint arXiv:2303.08774}, 2023.

\bibitem[Caron et~al.(2021)Caron, Touvron, Misra, J{\'e}gou, Mairal, Bojanowski, and Joulin]{dino}
Mathilde Caron, Hugo Touvron, Ishan Misra, Herv{\'e} J{\'e}gou, Julien Mairal, Piotr Bojanowski, and Armand Joulin.
\newblock Emerging properties in self-supervised vision transformers.
\newblock In \emph{Proceedings of the IEEE/CVF international conference on computer vision}, pages 9650--9660, 2021.

\bibitem[Chang et~al.(2023)Chang, Zhang, Barber, Maschinot, Lezama, Jiang, Yang, Murphy, Freeman, Rubinstein, et~al.]{muse}
Huiwen Chang, Han Zhang, Jarred Barber, AJ Maschinot, Jose Lezama, Lu Jiang, Ming-Hsuan Yang, Kevin Murphy, William~T Freeman, Michael Rubinstein, et~al.
\newblock Muse: Text-to-image generation via masked generative transformers.
\newblock \emph{Proceedings of the 40th International Conference on Machine Learning}, 202, 2023.

\bibitem[Dosovitskiy(2020)]{vit}
Alexey Dosovitskiy.
\newblock An image is worth 16x16 words: Transformers for image recognition at scale.
\newblock \emph{arXiv preprint arXiv:2010.11929}, 2020.

\bibitem[Gal et~al.(2022)Gal, Alaluf, Atzmon, Patashnik, Bermano, Chechik, and Cohen-Or]{textualinversion}
Rinon Gal, Yuval Alaluf, Yuval Atzmon, Or Patashnik, Amit~H Bermano, Gal Chechik, and Daniel Cohen-Or.
\newblock An image is worth one word: Personalizing text-to-image generation using textual inversion.
\newblock \emph{arXiv preprint arXiv:2208.01618}, 2022.

\bibitem[Grand et~al.(2018)Grand, Blank, Pereira, and Fedorenko]{projection_human}
Gabriel Grand, Idan~Asher Blank, Francisco Pereira, and Evelina Fedorenko.
\newblock Semantic projection: recovering human knowledge of multiple, distinct object features from word embeddings.
\newblock \emph{arXiv preprint arXiv:1802.01241}, 2018.

\bibitem[Guttenberg(2023)]{offsetnoise}
Nicholas Guttenberg.
\newblock Diffusion with offset noise.
\newblock \url{https://exampleblog.com/diffusion-offset-noise}, 2023.
\newblock Blog post.

\bibitem[Han et~al.(2023)Han, Li, Zhang, Milanfar, Metaxas, and Yang]{svdiff}
Ligong Han, Yinxiao Li, Han Zhang, Peyman Milanfar, Dimitris Metaxas, and Feng Yang.
\newblock Svdiff: Compact parameter space for diffusion fine-tuning.
\newblock In \emph{Proceedings of the IEEE/CVF International Conference on Computer Vision}, pages 7323--7334, 2023.

\bibitem[Hertz et~al.(2022)Hertz, Mokady, Tenenbaum, Aberman, Pritch, and Cohen-Or]{ptp}
Amir Hertz, Ron Mokady, Jay Tenenbaum, Kfir Aberman, Yael Pritch, and Daniel Cohen-Or.
\newblock Prompt-to-prompt image editing with cross attention control.
\newblock \emph{arXiv preprint arXiv:2208.01626}, 2022.

\bibitem[Hertz et~al.(2024)Hertz, Voynov, Fruchter, and Cohen-Or]{stylealigned}
Amir Hertz, Andrey Voynov, Shlomi Fruchter, and Daniel Cohen-Or.
\newblock Style aligned image generation via shared attention.
\newblock In \emph{Proceedings of the IEEE/CVF Conference on Computer Vision and Pattern Recognition}, pages 4775--4785, 2024.

\bibitem[Ho and Salimans(2022)]{cfg}
Jonathan Ho and Tim Salimans.
\newblock Classifier-free diffusion guidance.
\newblock \emph{arXiv preprint arXiv:2207.12598}, 2022.

\bibitem[Ho et~al.(2020)Ho, Jain, and Abbeel]{ddpm}
Jonathan Ho, Ajay Jain, and Pieter Abbeel.
\newblock Denoising diffusion probabilistic models.
\newblock \emph{Advances in neural information processing systems}, 33:\penalty0 6840--6851, 2020.

\bibitem[Houlsby et~al.(2019)Houlsby, Giurgiu, Jastrzebski, Morrone, De~Laroussilhe, Gesmundo, Attariyan, and Gelly]{peft}
Neil Houlsby, Andrei Giurgiu, Stanislaw Jastrzebski, Bruna Morrone, Quentin De~Laroussilhe, Andrea Gesmundo, Mona Attariyan, and Sylvain Gelly.
\newblock Parameter-efficient transfer learning for nlp.
\newblock In \emph{International conference on machine learning}, pages 2790--2799. PMLR, 2019.

\bibitem[Hu et~al.(2021)Hu, Shen, Wallis, Allen-Zhu, Li, Wang, Wang, and Chen]{lora}
Edward~J Hu, Yelong Shen, Phillip Wallis, Zeyuan Allen-Zhu, Yuanzhi Li, Shean Wang, Lu Wang, and Weizhu Chen.
\newblock Lora: Low-rank adaptation of large language models.
\newblock \emph{arXiv preprint arXiv:2106.09685}, 2021.

\bibitem[Jeong et~al.(2024)Jeong, Kim, Choi, Lee, and Uh]{vsp}
Jaeseok Jeong, Junho Kim, Yunjey Choi, Gayoung Lee, and Youngjung Uh.
\newblock Visual style prompting with swapping self-attention.
\newblock \emph{arXiv preprint arXiv:2402.12974}, 2024.

\bibitem[Kirstain et~al.(2023)Kirstain, Polyak, Singer, Matiana, Penna, and Levy]{pickapic}
Yuval Kirstain, Adam Polyak, Uriel Singer, Shahbuland Matiana, Joe Penna, and Omer Levy.
\newblock Pick-a-pic: An open dataset of user preferences for text-to-image generation.
\newblock \emph{Advances in Neural Information Processing Systems}, 36:\penalty0 36652--36663, 2023.

\bibitem[Klema and Laub(1980)]{svd}
Virginia Klema and Alan Laub.
\newblock The singular value decomposition: Its computation and some applications.
\newblock \emph{IEEE Transactions on automatic control}, 25\penalty0 (2):\penalty0 164--176, 1980.

\bibitem[Kumari et~al.(2023)Kumari, Zhang, Zhang, Shechtman, and Zhu]{customdiffusion}
Nupur Kumari, Bingliang Zhang, Richard Zhang, Eli Shechtman, and Jun-Yan Zhu.
\newblock Multi-concept customization of text-to-image diffusion.
\newblock In \emph{Proceedings of the IEEE/CVF Conference on Computer Vision and Pattern Recognition}, pages 1931--1941, 2023.

\bibitem[Lee et~al.(2024)Lee, Kwak, Sohn, and Shin]{dco}
Kyungmin Lee, Sangkyung Kwak, Kihyuk Sohn, and Jinwoo Shin.
\newblock Direct consistency optimization for compositional text-to-image personalization.
\newblock \emph{arXiv preprint arXiv:2402.12004}, 2024.

\bibitem[Lester et~al.(2021)Lester, Al-Rfou, and Constant]{peft2}
Brian Lester, Rami Al-Rfou, and Noah Constant.
\newblock The power of scale for parameter-efficient prompt tuning.
\newblock \emph{arXiv preprint arXiv:2104.08691}, 2021.

\bibitem[Li et~al.(2024)Li, van~de Weijer, Hu, Khan, Hou, Wang, and Yang]{getwhatyouwant}
Senmao Li, Joost van~de Weijer, Taihang Hu, Fahad~Shahbaz Khan, Qibin Hou, Yaxing Wang, and Jian Yang.
\newblock Get what you want, not what you don't: Image content suppression for text-to-image diffusion models.
\newblock \emph{arXiv preprint arXiv:2402.05375}, 2024.

\bibitem[Liu et~al.(2021)Liu, Ji, Fu, Tam, Du, Yang, and Tang]{peft3}
Xiao Liu, Kaixuan Ji, Yicheng Fu, Weng~Lam Tam, Zhengxiao Du, Zhilin Yang, and Jie Tang.
\newblock P-tuning v2: Prompt tuning can be comparable to fine-tuning universally across scales and tasks.
\newblock \emph{arXiv preprint arXiv:2110.07602}, 2021.

\bibitem[Liu et~al.(2023)Liu, Zheng, Du, Ding, Qian, Yang, and Tang]{peft1}
Xiao Liu, Yanan Zheng, Zhengxiao Du, Ming Ding, Yujie Qian, Zhilin Yang, and Jie Tang.
\newblock Gpt understands, too.
\newblock \emph{AI Open}, 2023.

\bibitem[Meng et~al.(2022)Meng, Bau, Andonian, and Belinkov]{locatinggpt}
Kevin Meng, David Bau, Alex Andonian, and Yonatan Belinkov.
\newblock Locating and editing factual associations in gpt.
\newblock \emph{Advances in Neural Information Processing Systems}, 35:\penalty0 17359--17372, 2022.

\bibitem[Miyake et~al.(2023)Miyake, Iohara, Saito, and Tanaka]{npi}
Daiki Miyake, Akihiro Iohara, Yu Saito, and Toshiyuki Tanaka.
\newblock Negative-prompt inversion: Fast image inversion for editing with text-guided diffusion models.
\newblock \emph{arXiv preprint arXiv:2305.16807}, 2023.

\bibitem[Mokady et~al.(2023)Mokady, Hertz, Aberman, Pritch, and Cohen-Or]{nti}
Ron Mokady, Amir Hertz, Kfir Aberman, Yael Pritch, and Daniel Cohen-Or.
\newblock Null-text inversion for editing real images using guided diffusion models.
\newblock In \emph{Proceedings of the IEEE/CVF Conference on Computer Vision and Pattern Recognition}, pages 6038--6047, 2023.

\bibitem[Oquab et~al.(2023)Oquab, Darcet, Moutakanni, Vo, Szafraniec, Khalidov, Fernandez, Haziza, Massa, El-Nouby, et~al.]{dinov2}
Maxime Oquab, Timoth{\'e}e Darcet, Th{\'e}o Moutakanni, Huy Vo, Marc Szafraniec, Vasil Khalidov, Pierre Fernandez, Daniel Haziza, Francisco Massa, Alaaeldin El-Nouby, et~al.
\newblock Dinov2: Learning robust visual features without supervision.
\newblock \emph{arXiv preprint arXiv:2304.07193}, 2023.

\bibitem[Ostashev et~al.(2024)Ostashev, Fang, Tulyakov, Aberman, et~al.]{moa}
Daniil Ostashev, Yuwei Fang, Sergey Tulyakov, Kfir Aberman, et~al.
\newblock Moa: Mixture-of-attention for subject-context disentanglement in personalized image generation.
\newblock \emph{arXiv preprint arXiv:2404.11565}, 2024.

\bibitem[Patil et~al.(2024)Patil, Berman, Rombach, and von Platen]{amused}
Suraj Patil, William Berman, Robin Rombach, and Patrick von Platen.
\newblock amused: An open muse reproduction.
\newblock \emph{arXiv preprint arXiv:2401.01808}, 2024.

\bibitem[Podell et~al.(2023)Podell, English, Lacey, Blattmann, Dockhorn, M{\"u}ller, Penna, and Rombach]{sdxl}
Dustin Podell, Zion English, Kyle Lacey, Andreas Blattmann, Tim Dockhorn, Jonas M{\"u}ller, Joe Penna, and Robin Rombach.
\newblock Sdxl: Improving latent diffusion models for high-resolution image synthesis.
\newblock \emph{arXiv preprint arXiv:2307.01952}, 2023.

\bibitem[Qin et~al.(2020)Qin, Hu, and Liu]{projection_cls}
Qi Qin, Wenpeng Hu, and Bing Liu.
\newblock Feature projection for improved text classification.
\newblock In \emph{Proceedings of the 58th Annual Meeting of the Association for Computational Linguistics}, pages 8161--8171, 2020.

\bibitem[Radford et~al.(2021)Radford, Kim, Hallacy, Ramesh, Goh, Agarwal, Sastry, Askell, Mishkin, Clark, et~al.]{clip}
Alec Radford, Jong~Wook Kim, Chris Hallacy, Aditya Ramesh, Gabriel Goh, Sandhini Agarwal, Girish Sastry, Amanda Askell, Pamela Mishkin, Jack Clark, et~al.
\newblock Learning transferable visual models from natural language supervision.
\newblock In \emph{International conference on machine learning}, pages 8748--8763. PMLR, 2021.

\bibitem[Rafailov et~al.(2024)Rafailov, Sharma, Mitchell, Manning, Ermon, and Finn]{dpo}
Rafael Rafailov, Archit Sharma, Eric Mitchell, Christopher~D Manning, Stefano Ermon, and Chelsea Finn.
\newblock Direct preference optimization: Your language model is secretly a reward model.
\newblock \emph{Advances in Neural Information Processing Systems}, 36, 2024.

\bibitem[Ramesh et~al.(2021)Ramesh, Pavlov, Goh, Gray, Voss, Radford, Chen, and Sutskever]{dalle}
Aditya Ramesh, Mikhail Pavlov, Gabriel Goh, Scott Gray, Chelsea Voss, Alec Radford, Mark Chen, and Ilya Sutskever.
\newblock Zero-shot text-to-image generation.
\newblock In \emph{International conference on machine learning}, pages 8821--8831. Pmlr, 2021.

\bibitem[Rombach et~al.(2022)Rombach, Blattmann, Lorenz, Esser, and Ommer]{stable_diffusion}
Robin Rombach, Andreas Blattmann, Dominik Lorenz, Patrick Esser, and Bj{\"o}rn Ommer.
\newblock High-resolution image synthesis with latent diffusion models.
\newblock In \emph{Proceedings of the IEEE/CVF conference on computer vision and pattern recognition}, pages 10684--10695, 2022.

\bibitem[Ronneberger et~al.(2015)Ronneberger, Fischer, and Brox]{unet}
Olaf Ronneberger, Philipp Fischer, and Thomas Brox.
\newblock U-net: Convolutional networks for biomedical image segmentation.
\newblock In \emph{Medical image computing and computer-assisted intervention--MICCAI 2015: 18th international conference, Munich, Germany, October 5-9, 2015, proceedings, part III 18}, pages 234--241. Springer, 2015.

\bibitem[Ruiz et~al.(2023)Ruiz, Li, Jampani, Pritch, Rubinstein, and Aberman]{dreambooth}
Nataniel Ruiz, Yuanzhen Li, Varun Jampani, Yael Pritch, Michael Rubinstein, and Kfir Aberman.
\newblock Dreambooth: Fine tuning text-to-image diffusion models for subject-driven generation.
\newblock In \emph{Proceedings of the IEEE/CVF conference on computer vision and pattern recognition}, pages 22500--22510, 2023.

\bibitem[Saharia et~al.(2022)Saharia, Chan, Saxena, Li, Whang, Denton, Ghasemipour, Gontijo~Lopes, Karagol~Ayan, Salimans, et~al.]{imagen}
Chitwan Saharia, William Chan, Saurabh Saxena, Lala Li, Jay Whang, Emily~L Denton, Kamyar Ghasemipour, Raphael Gontijo~Lopes, Burcu Karagol~Ayan, Tim Salimans, et~al.
\newblock Photorealistic text-to-image diffusion models with deep language understanding.
\newblock \emph{Advances in neural information processing systems}, 35:\penalty0 36479--36494, 2022.

\bibitem[Schmidt(1907)]{schmidt}
Erhard Schmidt.
\newblock Zur theorie der linearen und nichtlinearen integralgleichungen.
\newblock \emph{Mathematische Annalen}, 63\penalty0 (4):\penalty0 433--476, 1907.

\bibitem[Shah et~al.(2025)Shah, Ruiz, Cole, Lu, Lazebnik, Li, and Jampani]{ziplora}
Viraj Shah, Nataniel Ruiz, Forrester Cole, Erika Lu, Svetlana Lazebnik, Yuanzhen Li, and Varun Jampani.
\newblock Ziplora: Any subject in any style by effectively merging loras.
\newblock In \emph{European Conference on Computer Vision}, pages 422--438. Springer, 2025.

\bibitem[Shing(2023)]{ziprora_unofficial}
Makoto Shing.
\newblock Ziplora-pytorch.
\newblock \url{https://github.com/mkshing/ziplora-pytorch}, 2023.
\newblock GitHub repository.

\bibitem[Sohn et~al.(2023)Sohn, Ruiz, Lee, Chin, Blok, Chang, Barber, Jiang, Entis, Li, et~al.]{styledrop}
Kihyuk Sohn, Nataniel Ruiz, Kimin Lee, Daniel~Castro Chin, Irina Blok, Huiwen Chang, Jarred Barber, Lu Jiang, Glenn Entis, Yuanzhen Li, et~al.
\newblock Styledrop: Text-to-image generation in any style.
\newblock \emph{arXiv preprint arXiv:2306.00983}, 2023.

\bibitem[Somepalli et~al.(2024)Somepalli, Gupta, Gupta, Palta, Goldblum, Geiping, Shrivastava, and Goldstein]{csd}
Gowthami Somepalli, Anubhav Gupta, Kamal Gupta, Shramay Palta, Micah Goldblum, Jonas Geiping, Abhinav Shrivastava, and Tom Goldstein.
\newblock Measuring style similarity in diffusion models.
\newblock \emph{arXiv preprint arXiv:2404.01292}, 2024.

\bibitem[Song et~al.(2020)Song, Meng, and Ermon]{ddim}
Jiaming Song, Chenlin Meng, and Stefano Ermon.
\newblock Denoising diffusion implicit models.
\newblock \emph{arXiv preprint arXiv:2010.02502}, 2020.

\bibitem[Tewel et~al.(2023)Tewel, Gal, Chechik, and Atzmon]{perfusion}
Yoad Tewel, Rinon Gal, Gal Chechik, and Yuval Atzmon.
\newblock Key-locked rank one editing for text-to-image personalization.
\newblock In \emph{ACM SIGGRAPH 2023 Conference Proceedings}, pages 1--11, 2023.

\bibitem[Wu et~al.(2023)Wu, Hao, Sun, Chen, Zhu, Zhao, and Li]{hpsv2}
Xiaoshi Wu, Yiming Hao, Keqiang Sun, Yixiong Chen, Feng Zhu, Rui Zhao, and Hongsheng Li.
\newblock Human preference score v2: A solid benchmark for evaluating human preferences of text-to-image synthesis.
\newblock \emph{arXiv preprint arXiv:2306.09341}, 2023.

\bibitem[Xiao et~al.(2023)Xiao, Yin, Freeman, Durand, and Han]{fastcomposer}
Guangxuan Xiao, Tianwei Yin, William~T Freeman, Fr{\'e}do Durand, and Song Han.
\newblock Fastcomposer: Tuning-free multi-subject image generation with localized attention.
\newblock \emph{arXiv preprint arXiv:2305.10431}, 2023.

\bibitem[Xu et~al.(2024)Xu, Tang, Cao, Zhang, Deussen, Dong, Li, and Lee]{breakformake}
Yu Xu, Fan Tang, Juan Cao, Yuxin Zhang, Oliver Deussen, Weiming Dong, Jintao Li, and Tong-Yee Lee.
\newblock Break-for-make: Modular low-rank adaptations for composable content-style customization.
\newblock \emph{arXiv preprint arXiv:2403.19456}, 2024.

\bibitem[Ye et~al.(2023)Ye, Zhang, Liu, Han, and Yang]{ipadapter}
Hu Ye, Jun Zhang, Sibo Liu, Xiao Han, and Wei Yang.
\newblock Ip-adapter: Text compatible image prompt adapter for text-to-image diffusion models.
\newblock \emph{arXiv preprint arXiv:2308.06721}, 2023.

\bibitem[Zeng et~al.(2024)Zeng, Yan, Zhu, Chen, Chu, Zhao, and Yang]{infusion}
Weili Zeng, Yichao Yan, Qi Zhu, Zhuo Chen, Pengzhi Chu, Weiming Zhao, and Xiaokang Yang.
\newblock Infusion: Preventing customized text-to-image diffusion from overfitting.
\newblock \emph{arXiv preprint arXiv:2404.14007}, 2024.

\bibitem[Zhang et~al.(2023)Zhang, Rao, and Agrawala]{controlnet}
Lvmin Zhang, Anyi Rao, and Maneesh Agrawala.
\newblock Adding conditional control to text-to-image diffusion models.
\newblock In \emph{Proceedings of the IEEE/CVF International Conference on Computer Vision}, pages 3836--3847, 2023.

\bibitem[Zhuang et~al.(2024)Zhuang, Hu, and Gao]{magnet}
Chenyi Zhuang, Ying Hu, and Pan Gao.
\newblock Magnet: We never know how text-to-image diffusion models work, until we learn how vision-language models function.
\newblock \emph{arXiv preprint arXiv:2409.19967}, 2024.

\end{thebibliography}
}

\clearpage
\setcounter{page}{1}
\maketitlesupplementary

\setcounter{table}{0}
\renewcommand{\thetable}{A-\arabic{table}}
\setcounter{figure}{0}
\renewcommand{\thefigure}{A-\arabic{figure}}
\setcounter{equation}{0}
\renewcommand{\theequation}{A-\arabic{equation}}
\setcounter{section}{0}
\renewcommand{\thesection}{A-\arabic{section}}

\section{Additional experiments}

\subsection{Additional ablation study}

\begin{figure}[t]
    \centering
    \includegraphics[width=1.0\linewidth]{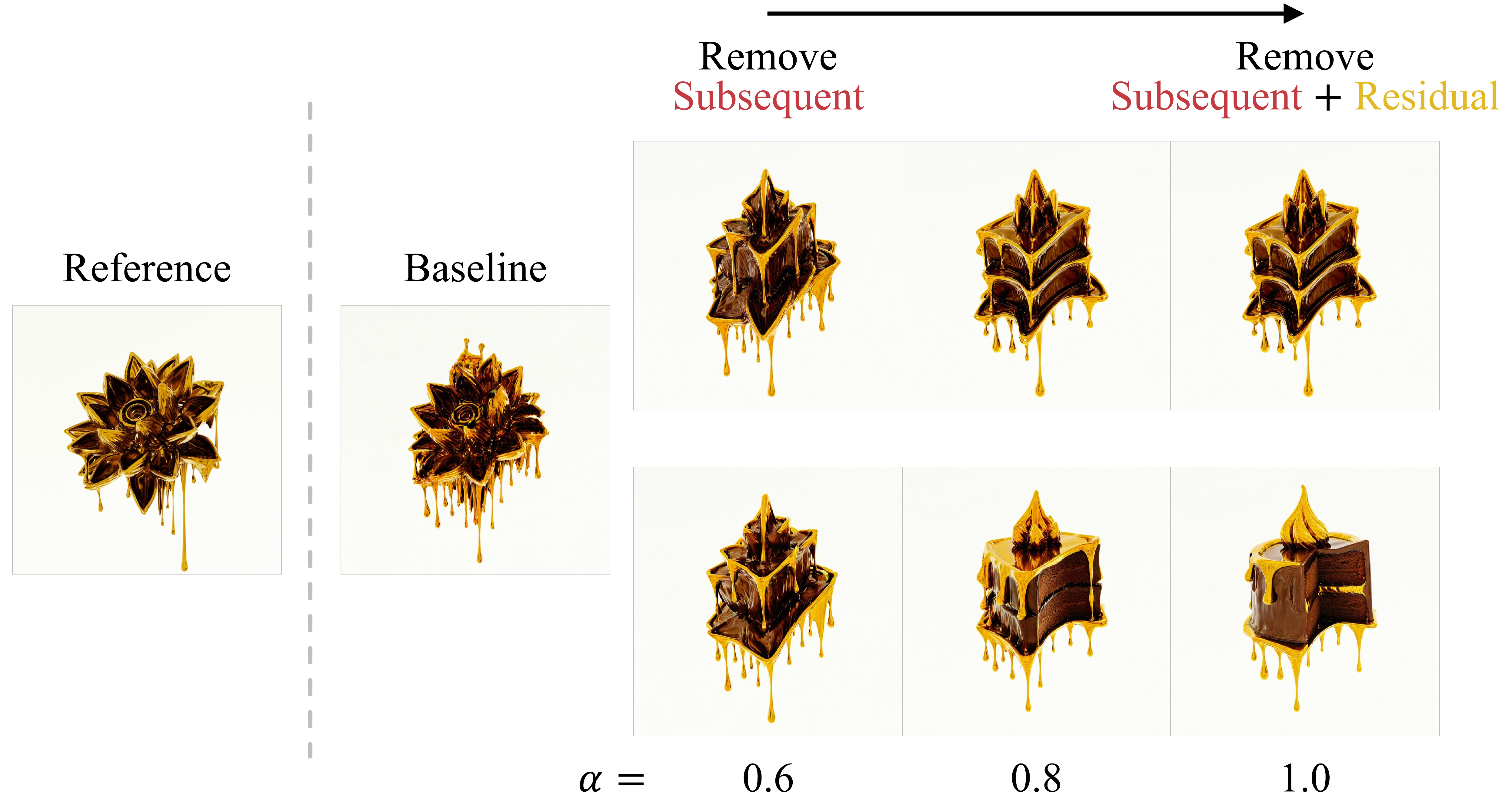}
    \caption{It is more effective to explicitly control the degree of separation from the unwanted embedding (bottom) rather than suppressing the embedding in an ambiguous manner (top).}
    \label{fig:additional_suppression_compare}
\end{figure}

In \cref{sub_sec:3_2}, we explore various approaches to suppressing the word token embedding $X_\textnormal{w}$. Among these, the method of suppressing subsequent components is indeterminate due to the ambiguity in singular value indexing. The top-left part of \cref{fig:additional_suppression_compare} shows the results of gradually increasing the index range of subsequent components to suppress $X_\textnormal{w}$, eventually removing all residual components. As the singular values corresponding to $X_\textnormal{w}$ are progressively removed, overfitting decreases, and the generated image better matches the given target (\eg, a chocolate cake). However, the results exhibit visual distortions.

In contrast, the bottom-left part of the figure demonstrates our method, where adjusting the parameter $\alpha$ explicitly controls the degree of separation from the $X_\textnormal{w}$ subspace. This adjustment ensures the target object is represented accurately and without distortion, highlighting the effectiveness of our approach.

\subsection{Additional results}

\cref{tab:comparison} presents the comprehensive quantitative results for personalization, stylization, and content-style mixing.

\begin{figure}[t]
    \centering
    \vspace{0.5em}
    \includegraphics[width=1.0\linewidth]{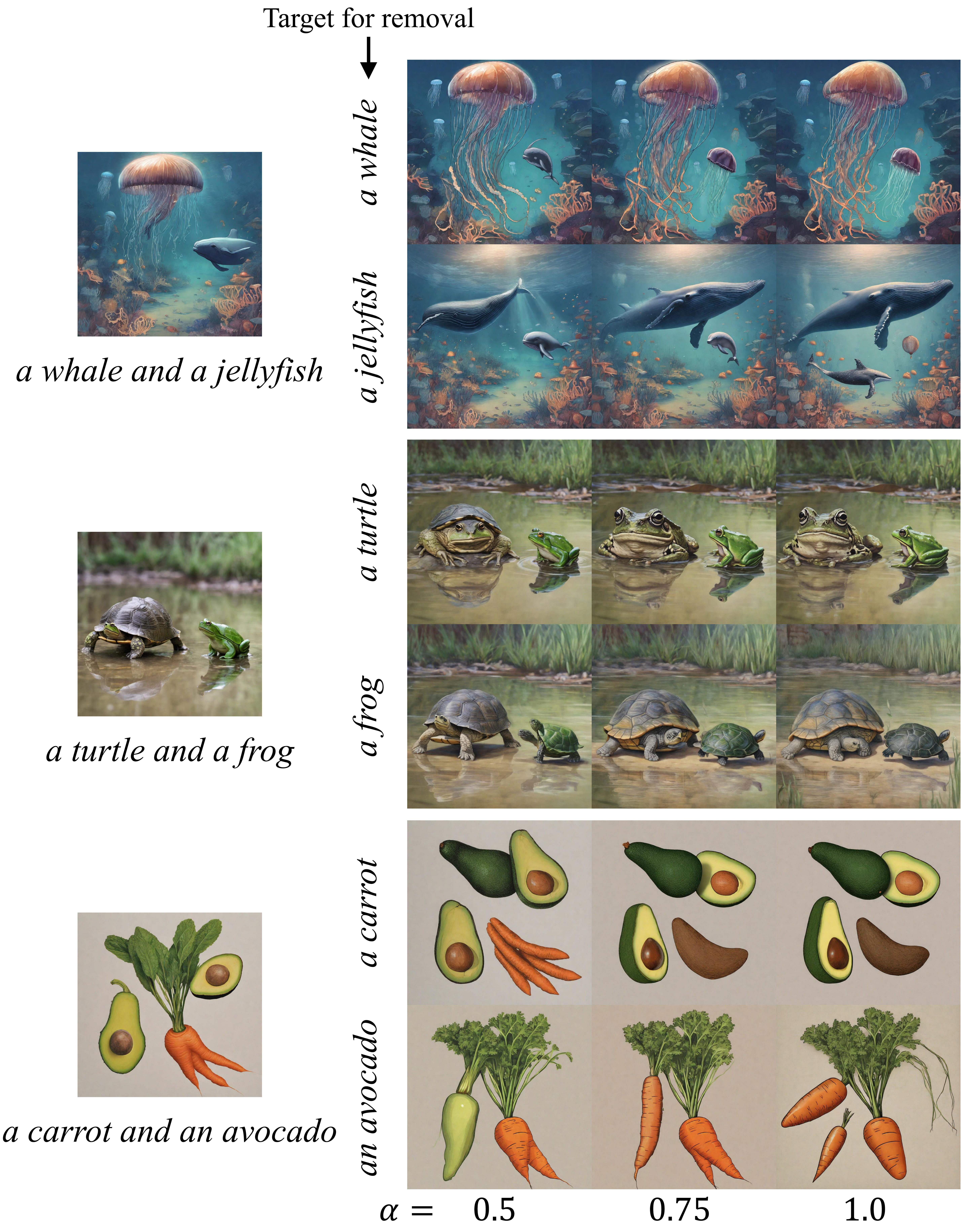}
    \caption{Image synthesis results generated using projected text embedding. From left to right, $\alpha$ is 0.5, 0.75, and 1.0.}
    \label{fig:vanila_sdxl}
\end{figure}

\noindent \textbf{T2I synthesis using text embedding projection.} \cref{fig:vanila_sdxl} shows results of text embedding modification. image generation results after removing the components of the original text embedding that belong to the embedding space of the target for removal. Modifications utilizing orthogonality in the text embedding space can effectively adjust the image generation trajectory. As future work, this embedding semantics modification technique could be combined with attention map manipulation methods~\cite{ptp, nti, npi} to enable more elaborate image editing.

\noindent \textbf{Contextualized generation of content-style mixing.} For the content-style mixing task, a specific subject can be depicted in a specific style with additional descriptive expressions. \cref{fig:contextualized_merge} shows examples where the subject is combined with other descriptions in the given style.

\noindent \textbf{Additional synthesized images.} \cref{fig:additional_personalization_results}, \cref{fig:additional_stylization_results}, and \cref{fig:additional_merge_results} respectively present additional results for personalization, stylization, and content-style mixing. Our method addresses prompt misalignment and content leakage issues found in DreamBooth.

\noindent \textbf{Controlling $\alpha$.} \cref{fig:additional_ablation} illustrates how the generated images change with adjustments to the projection intensity parameter $\alpha$. \cref{fig:additional_ablation} (a) shows the prevention of overfitting, while \cref{fig:additional_ablation} (b) illustrates the control over fine visual details.

\begin{figure}[t]
    \centering
    \includegraphics[width=0.75\linewidth]{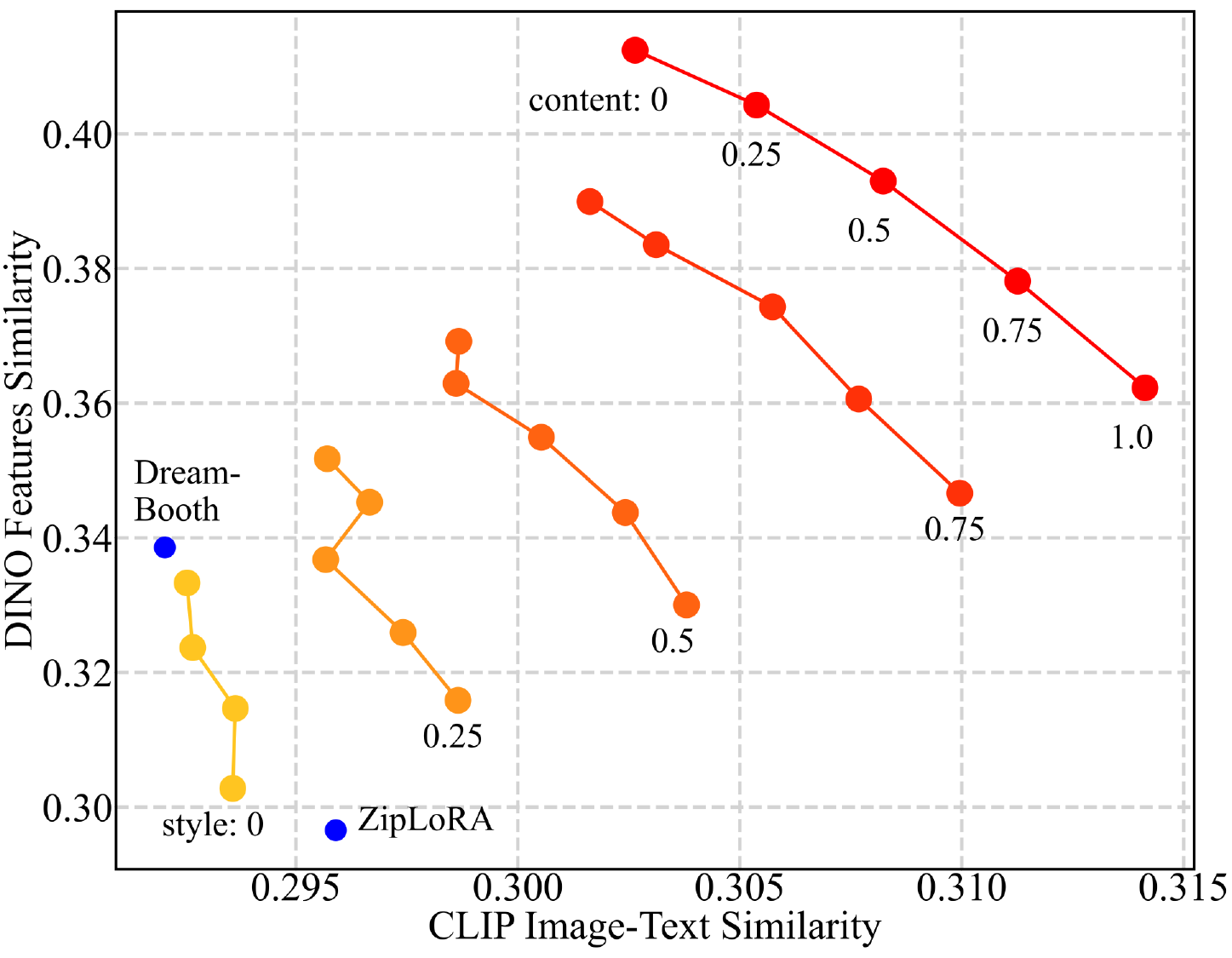}
    \caption{Quantitative content-style mixing comparison. In DECOR, The labels indicate the value of $\alpha$ of the text embeddings input to the content or style LoRA layers. An $\alpha=0$ indicates that the original embedding without projection was used. There is a trade-off depending on the text embeddings for each content and style LoRA layers.}
    \label{fig:sup1_mixing_score}
\end{figure}

\noindent \textbf{Quantitative comparison of content-style mixing.} \cref{fig:sup1_mixing_score} shows the quantitative results of content-style mixing for DECOR and the comparison methods. We evaluate a total of 25 combinations by inputting different embeddings, including the original embedding and projected embeddings with varying $\alpha$, into the content and style LoRA layers. DreamBooth corresponds to the combination where the original text embedding is used for both content and style LoRA, while the remaining 24 combinations are DECOR. When using the projected embedding with $\alpha=1.0$ as input for the style LoRA, DECOR shows the best performance compared to other cases. Additionally, the quantitative results indicate that changes in the text embedding for the style LoRA have a greater impact on the scores than changes in the content LoRA. This suggests that stylization has a greater impact on overall image features and text-image fidelity than personalization.

\begin{table}[t]
\centering
\begin{tabular}{@{}lcc@{}}
\toprule
                         & \textbf{CLIP t-i sim.} & \textbf{DINO sim.} \\ \midrule
Personalization          &               &             \\ \midrule
DreamBooth               & 0.285 $\pm$ 0.052       & \textbf{0.514 $\pm$ 0.183}     \\
Custom Diffusion         & 0.303 $\pm$ 0.042       & 0.444 $\pm$ 0.180     \\
Textual Inversion        & 0.301 $\pm$ 0.043       & 0.310 $\pm$ 0.141     \\
DECOR (Ours)             & \textbf{0.308 $\pm$ 0.038}       & 0.439 $\pm$ 0.172     \\ \midrule
Stylization              &               &             \\ \midrule
DreamBooth               & 0.285 $\pm$ 0.054       & 0.556 $\pm$ 0.185     \\
StyleDrop                & 0.291 $\pm$ 0.053       & 0.395 $\pm$ 0.144     \\
IP-Adapter               & 0.287 $\pm$ 0.053       & 0.556 $\pm$ 0.184     \\
StyleAligned             & 0.274 $\pm$ 0.049       & \textbf{0.573 $\pm$ 0.155}     \\
Visual Style Prompting   & 0.308 $\pm$ 0.041       & 0.384 $\pm$ 0.144     \\
DECOR (Ours)             & \textbf{0.310 $\pm$ 0.042}       & 0.472 $\pm$ 0.159     \\ \midrule
Content-style Mixing     &               &             \\ \midrule
DreamBooth               & 0.292 $\pm$ 0.037       & 0.339 $\pm$ 0.139     \\
ZipLoRA                  & 0.296 $\pm$ 0.041       & 0.297 $\pm$ 0.139     \\
DECOR (Ours)             & \textbf{0.305 $\pm$ 0.033}       & \textbf{0.404 $\pm$ 0.150}      \\ \bottomrule
\end{tabular}
\caption{Quantitative comparison of personalization, stylization, and content-style mixing. CLIP t-i sim. refers to CLIP text-image similarity, and DINO sim. refers to DINO feature similarity. For personalization and stylization, $\alpha=0.8$ is used. For content-style mixing, $\alpha=0.25$ is used for content LoRA and $\alpha=1.0$ for style LoRA.}
\label{tab:comparison}
\end{table}

\noindent \textbf{Comparison using preference models.} Recent studies have focused on predicting human preferences for text-image pairs to capture subtle preference distributions that cannot be identified using traditional evaluation metrics. PickScore~\cite{pickapic} is an evaluation model to assess the compatibility between text prompts and generated images. Human Preference Score v2~\cite{hpsv2} (HPS v2) calculates scores based on text-image alignment and aesthetic quality. These preference scoring models were trained based on the CLIP model and optimized using KL-divergence minimization to fit human preference distributions.

\begin{figure}[t]
    \centering
    \vspace{1em}
    \includegraphics[width=0.8\linewidth]{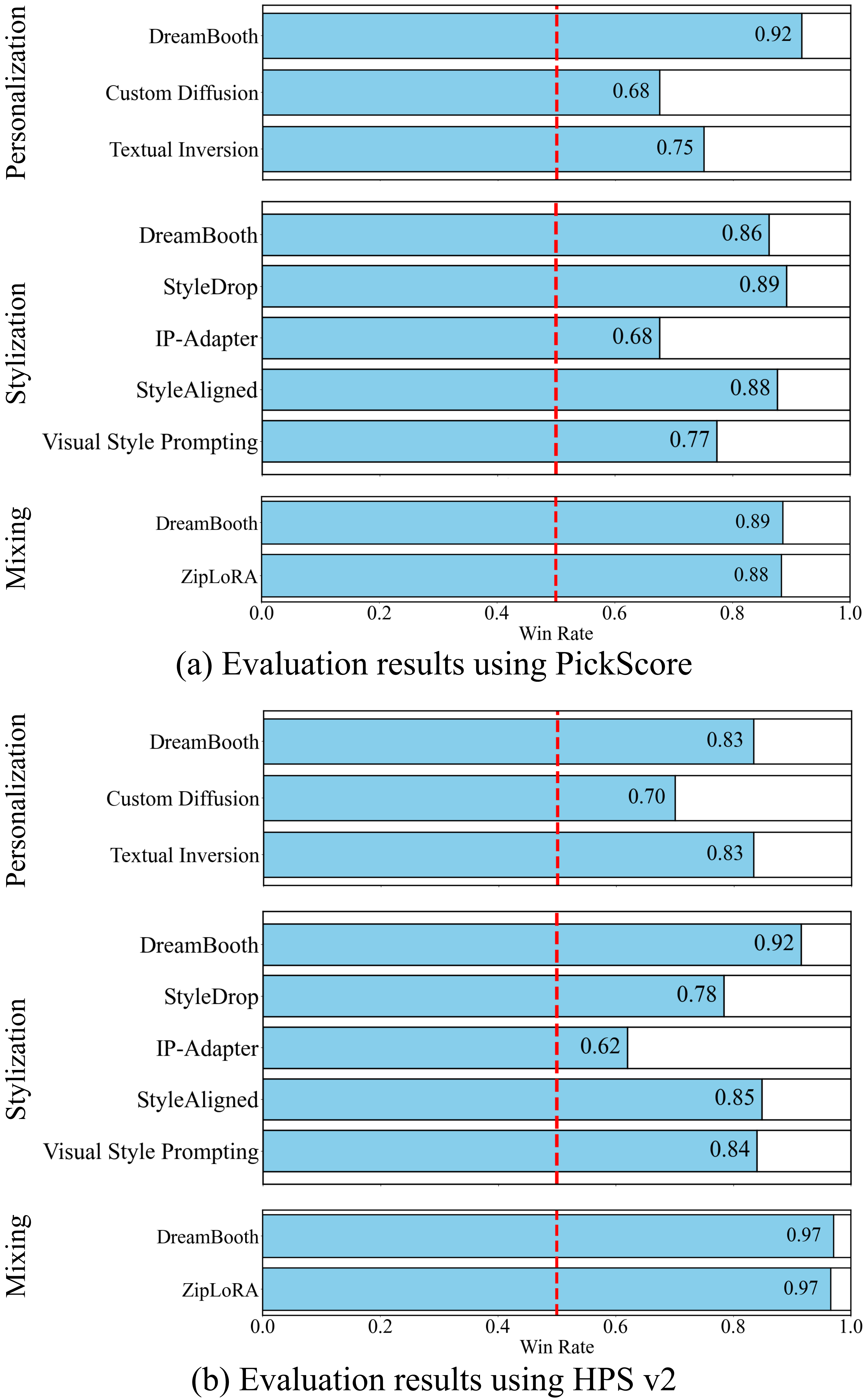}
    \caption{Quantitative results using human preference scoring models. A win rate exceeding the central red line (0.5) indicates that our method outperforms each comparative method.}
    \label{fig:preference_score}
\end{figure}

\begin{figure*}[t]
    \centering
    \includegraphics[width=1.0\linewidth]{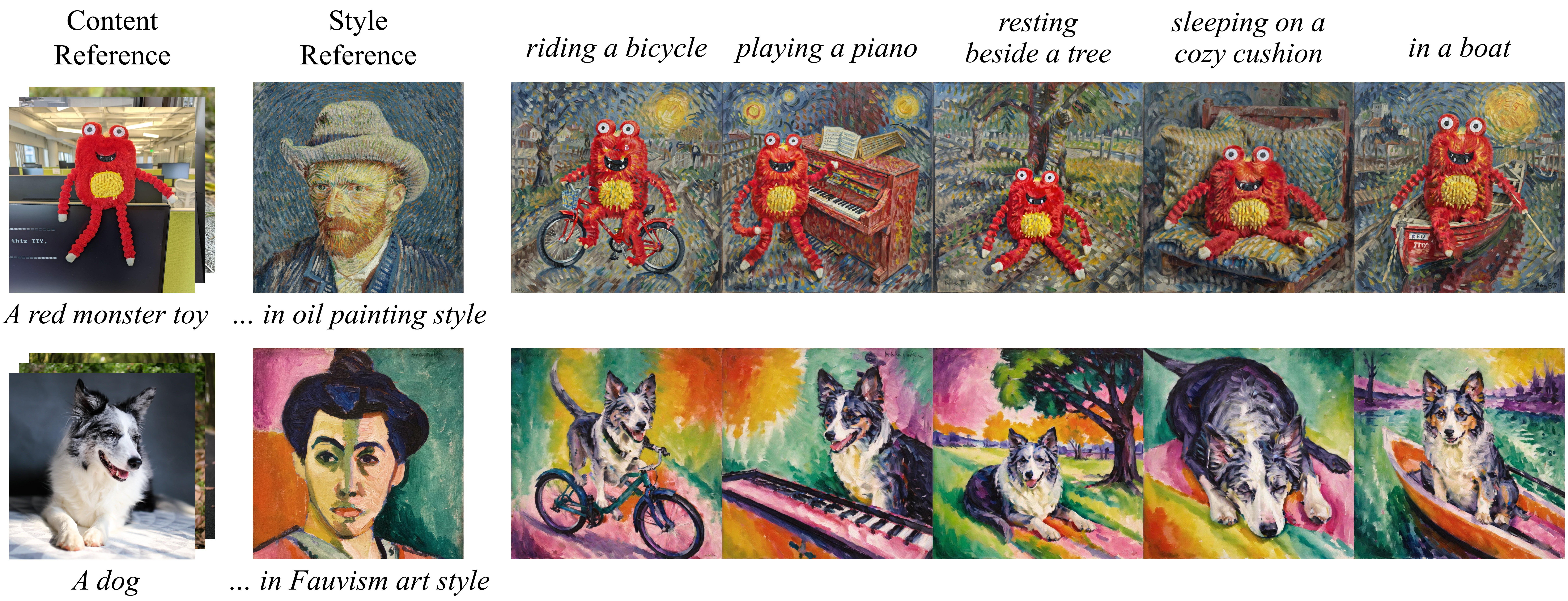}
    \caption{Contextualized mixing results. The subject is effectively represented in the given style while combining with various additional descriptions. Content LoRA uses $\alpha=0.25$, and style LoRA uses $\alpha=0.75$.}
    \label{fig:contextualized_merge}
\end{figure*}

\cref{fig:preference_score} shows the quantitative evaluation results of personalization and stylization using the two models. The win rate is calculated based on whether the model assigns a higher score to an image generated by our method compared to another image generated from the same prompt. As shown in the plot, our method achieves better performance in all cases. This demonstrates that our method outperforms existing approaches based on human aesthetic preferences and prompt alignment, which cannot be fully captured by conventional metrics such as CLIP or DINO similarity scores.

\subsection{Experimental details}

\begin{figure}[t]
    \centering
    \includegraphics[width=0.9\linewidth]{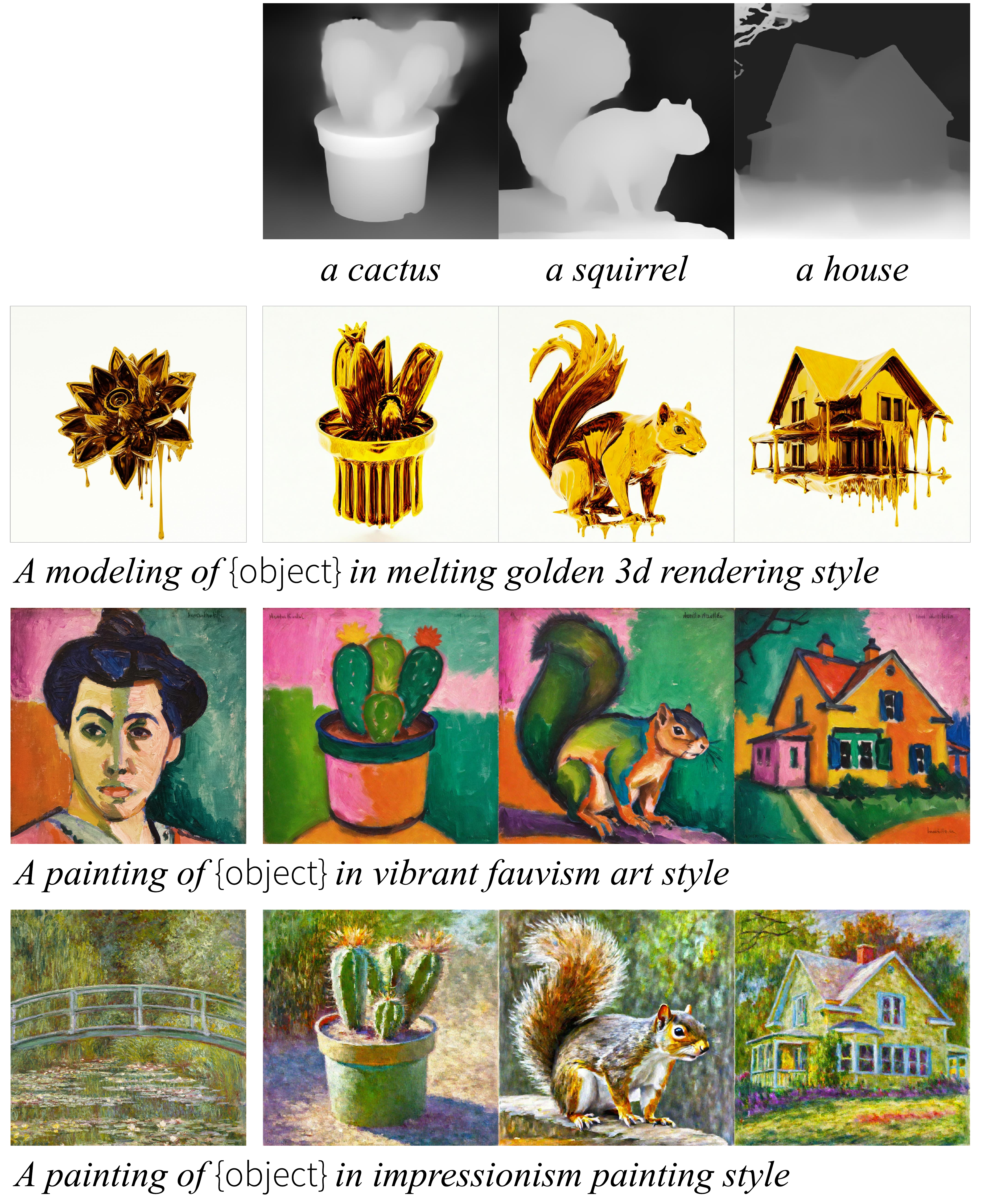}
    \caption{Results combined with ControlNet. $\alpha$ is set to 0.8.}
    \label{fig:sup2_controlnet}
\end{figure}

\noindent \textbf{Training details.} For content LoRA tuning, we created detailed captions for each reference image and set the main category word as a tunable special token (\eg, ``a \verb|<|dog\verb|>|''), following ~\cite{dco}. The LoRA rank was set to 32, with a learning rate of 5e-5 for the LoRA layer, 5e-6 for the special token embedding, a batch size of 1, and training conducted for 1,000 steps. For style LoRA tuning, no special token was used. The LoRA rank was set to 64, with a learning rate of 5e-5, a batch size of 1, and training conducted for 1000 steps. We set the denoising timestep to $T=50$ in the experiment framework and used the DDIM sampler~\cite{ddim}. We set the classifier-free guidance~\cite{cfg} (CFG) scale to 7.5, and the LoRA adapter scale to 1.0. Note that our method modifies the text embedding before inference, with an execution time difference of under a second compared to the primary baseline, DreamBooth.

\noindent \textbf{Evaluation templates.} For the geometric analysis of text embeddings in \cref{fig:fig2_sv_and_emb_analysis}, we used 20 text sentences of similar length. To demonstrate the common characteristics of prompt embeddings, we selected the sentences on diverse topics using the GPT-4o model ({\fontfamily{lmtt}\selectfont gpt-4o-2024-08-06})~\cite{gpt4}, as shown in \cref{fig:template} (a). \cref{fig:template} (b) and (c) also show example templates for personalization and stylization evaluation.

\subsection{Integration with other methodologies} 

\noindent \textbf{Combined with ControlNet.} Research has been conducted on incorporating various conditions into T2I generation models. ControlNet~\cite{controlnet} enables diffusion models to integrate visual conditions alongside text prompts by attaching additional trained layers to the model. \cref{fig:sup2_controlnet} shows stylization results combining our method with depthmap ControlNet. The results show that the model effectively aligns with the given depth information and text prompt, accurately representing each style.
 
\noindent \textbf{Combined with DCO loss.} Direct consistency optimization~\cite{dco} (DCO) introduces a new approach to fine-tuning diffusion models. It is inspired by direct preference optimization~\cite{dpo} (DPO), which demonstrated that policy model can be trained using preference datasets without the need for a reward model in language models domain. DCO adopts the loss from DPO and refines it for the sequential denoising process of diffusion models. It introduces a loss function that reduces the noise prediction error of the fine-tuned model smaller than that of the pre-trained model.

\begin{figure}[t]
    \centering
    \includegraphics[width=0.95\linewidth]{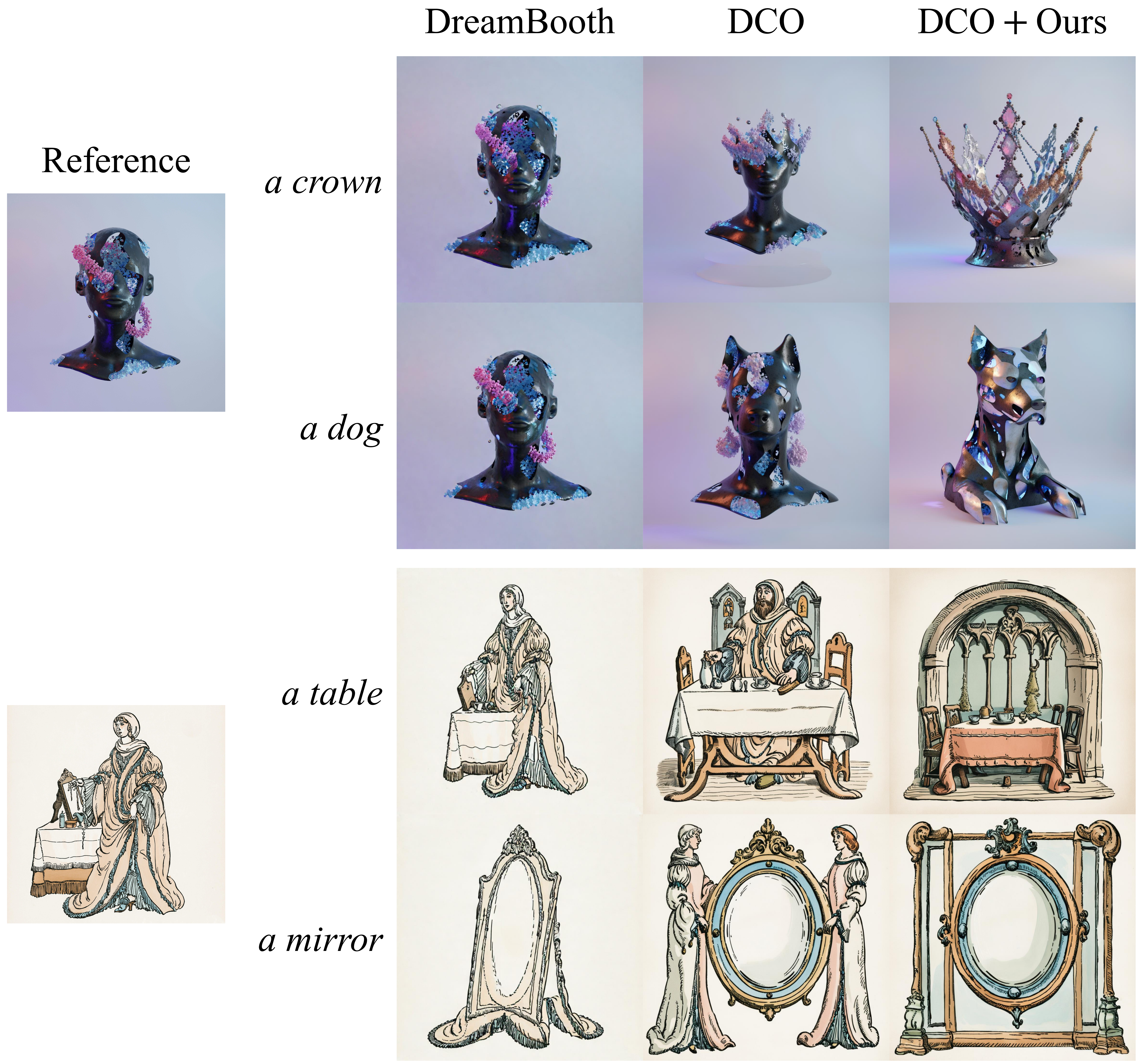}
    \caption{Results combined with DCO loss. $\alpha$ is set to 0.8 for DCO+Ours.}
    \label{fig:sup3_dco}
\end{figure}

Our method, DECOR, can integrate with this alternative loss function, as it adjusts input text embeddings during inference. \cref{fig:sup3_dco} shows results using the original loss (DreamBooth), results using only the DCO loss, and results combining the DCO loss with our method. Although training LoRA layers with the DCO loss significantly reduces overfitting compared to DreamBooth, the incorporation of our method further enables a more faithful synthesis of the desired target. These results demonstrate the integrability and extensibility of our method with other approaches.

\section{Limitation}

\begin{figure}[t]
    \centering
    \vspace{1em}
    \includegraphics[width=0.95\linewidth]{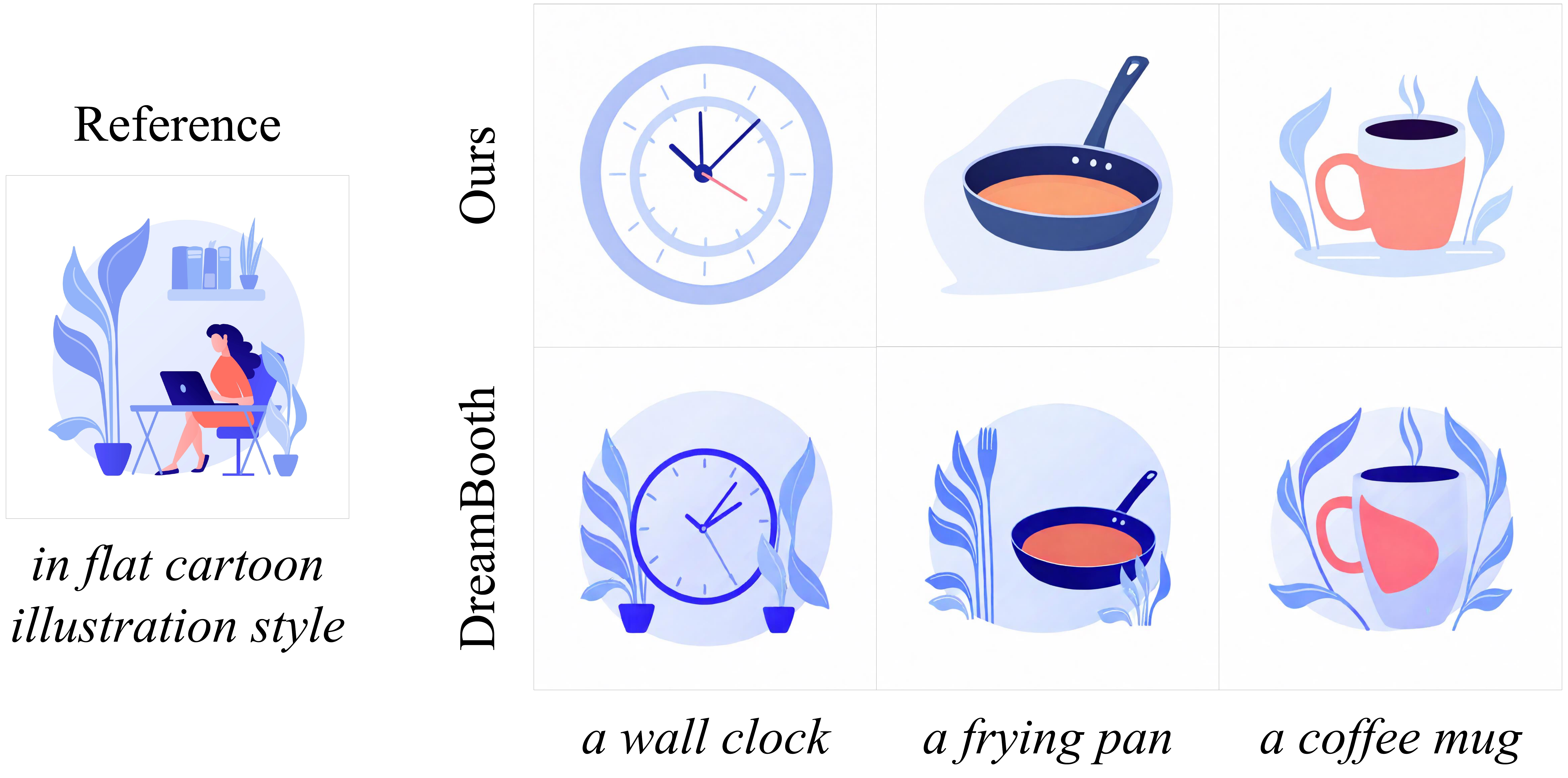}
    \caption{Limitations. In DECOR, deviating from the given style can lead to the loss of detailed stylistic characteristics. $\alpha$ is set to 1.0.}
    \label{fig:limitation}
\end{figure}

\begin{figure*}[ht]
    \centering
    \includegraphics[width=0.8\linewidth]{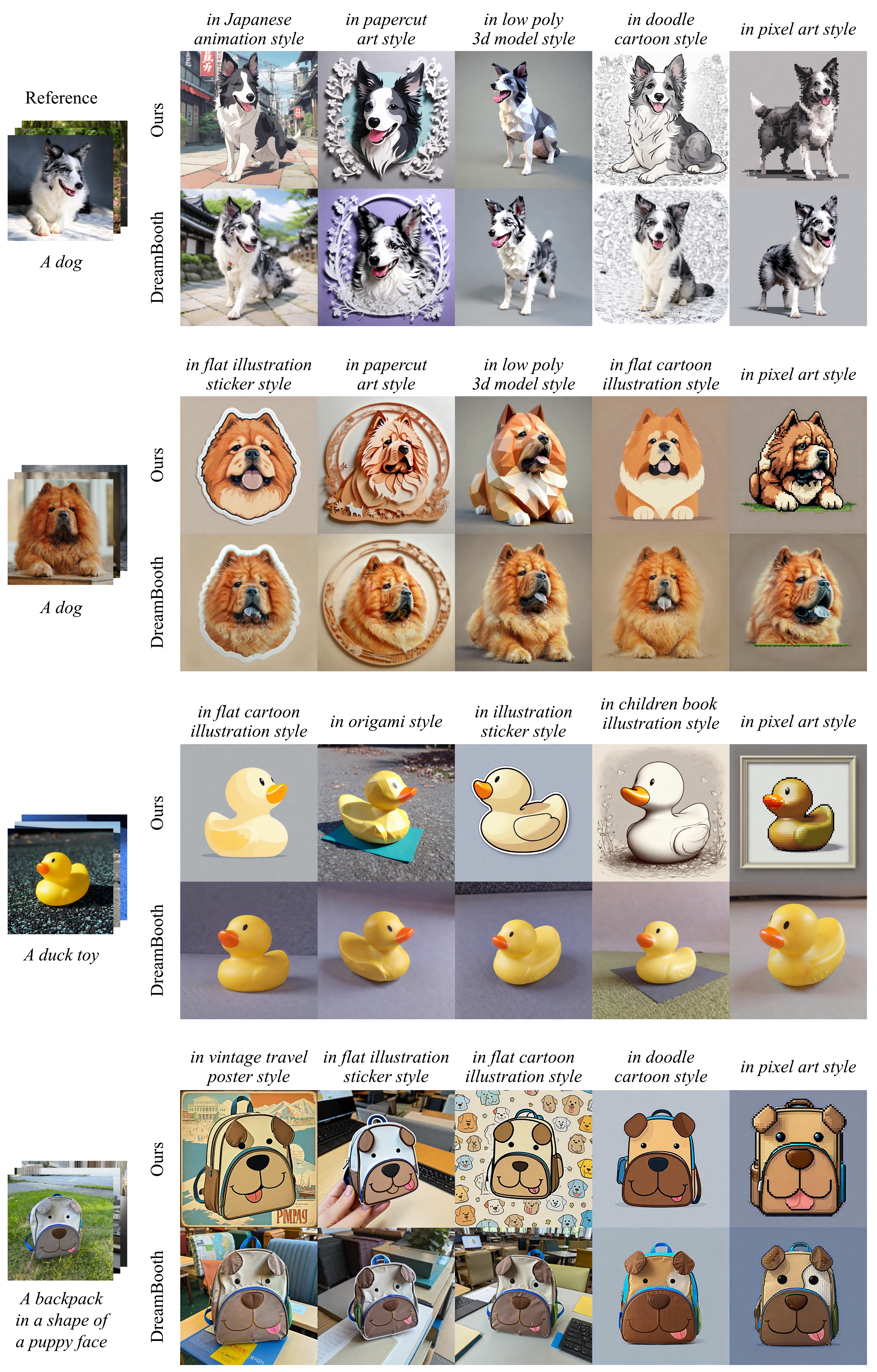}
    \caption{Additional personalization results. $\alpha$ is set to 0.8.}
    \label{fig:additional_personalization_results}
\end{figure*}

\begin{figure*}[ht]
    \centering
    \includegraphics[width=0.8\linewidth]{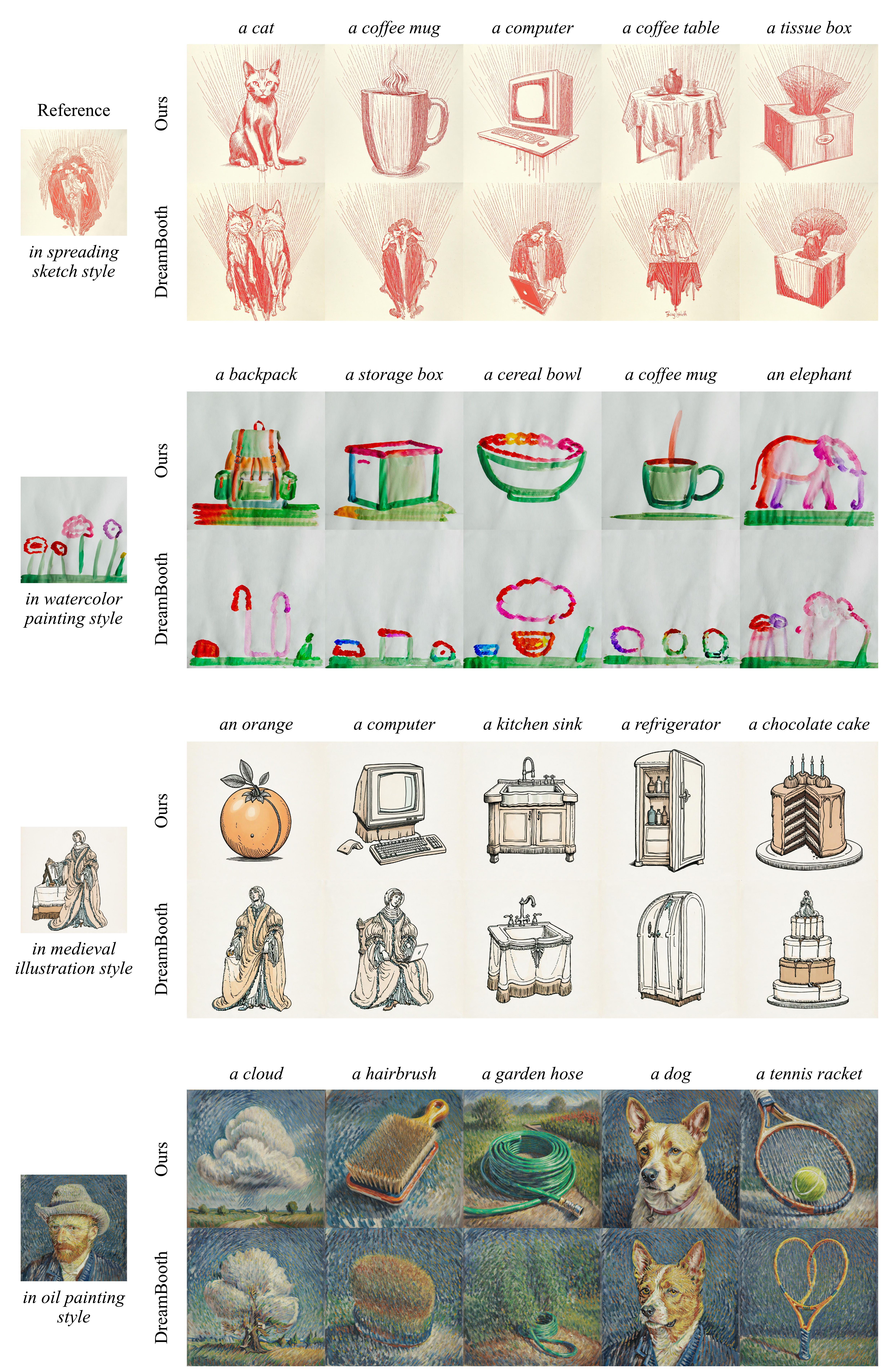}
    \caption{Additional stylization results. $\alpha$ is set to 0.8.}
    \label{fig:additional_stylization_results}
\end{figure*}

\begin{figure*}[ht]
    \centering
    \includegraphics[width=0.77\linewidth]{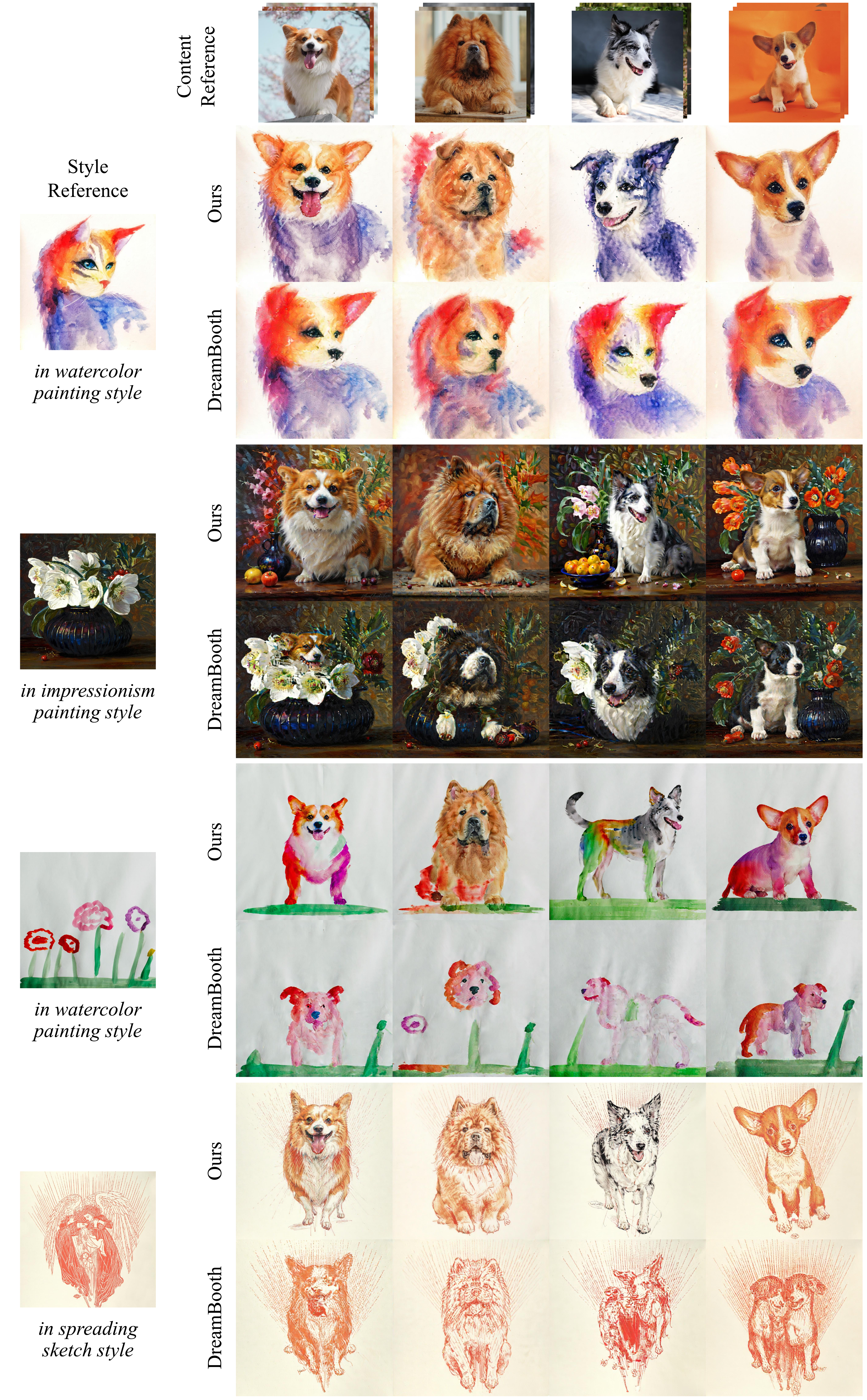}
    \caption{Additional content-style mixing results. $\alpha$ is set to 0.25 for content LoRA and varies from 0.5 to 1.0 for style LoRA.}
    \label{fig:additional_merge_results}
\end{figure*}

\begin{figure*}[ht]
    \centering
    \includegraphics[width=0.9\linewidth]{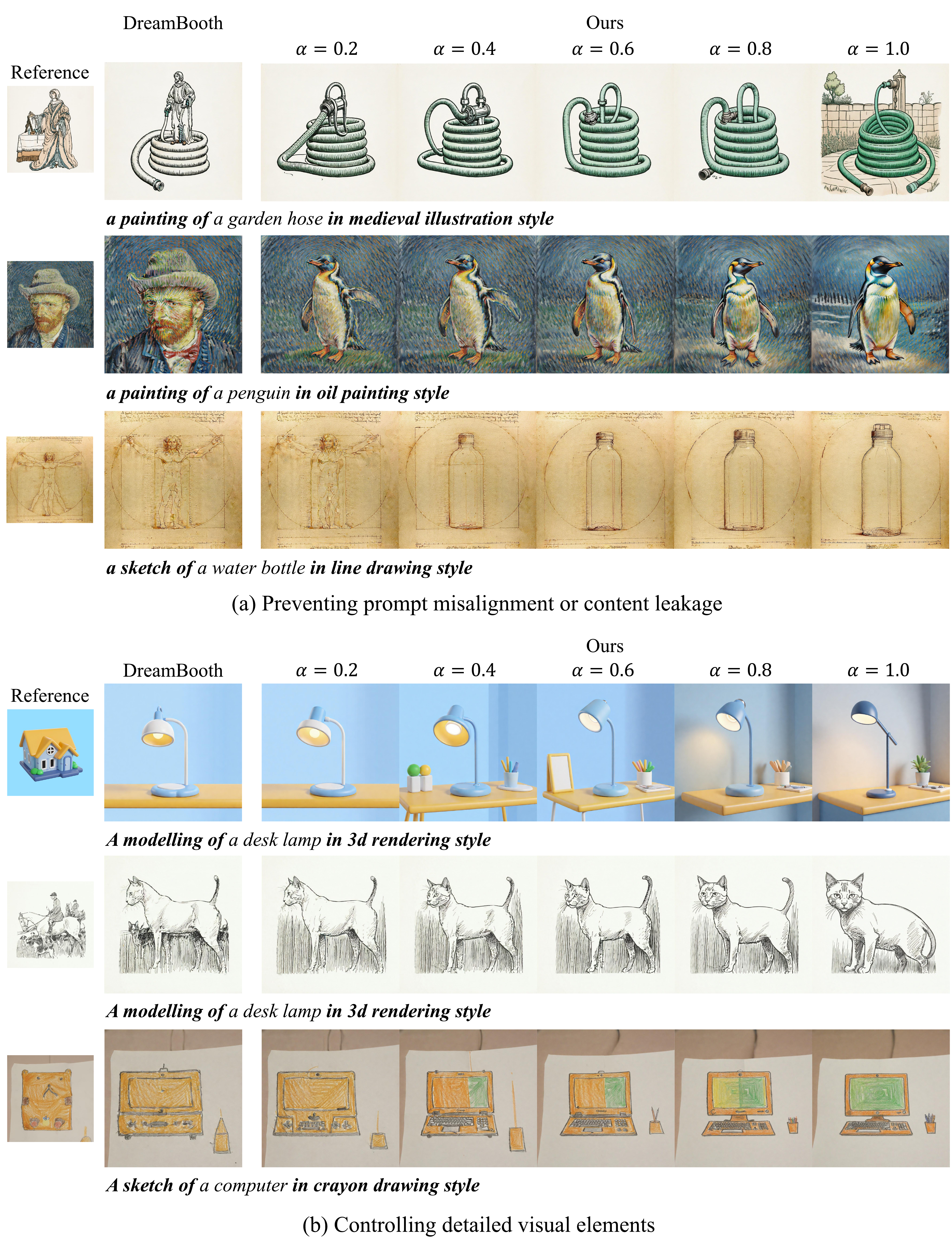}
    \caption{Additional results for controlling $\alpha$.}
    \label{fig:additional_ablation}
\end{figure*}

\begin{figure*}[ht]
    \centering
    \includegraphics[width=1.0\linewidth]{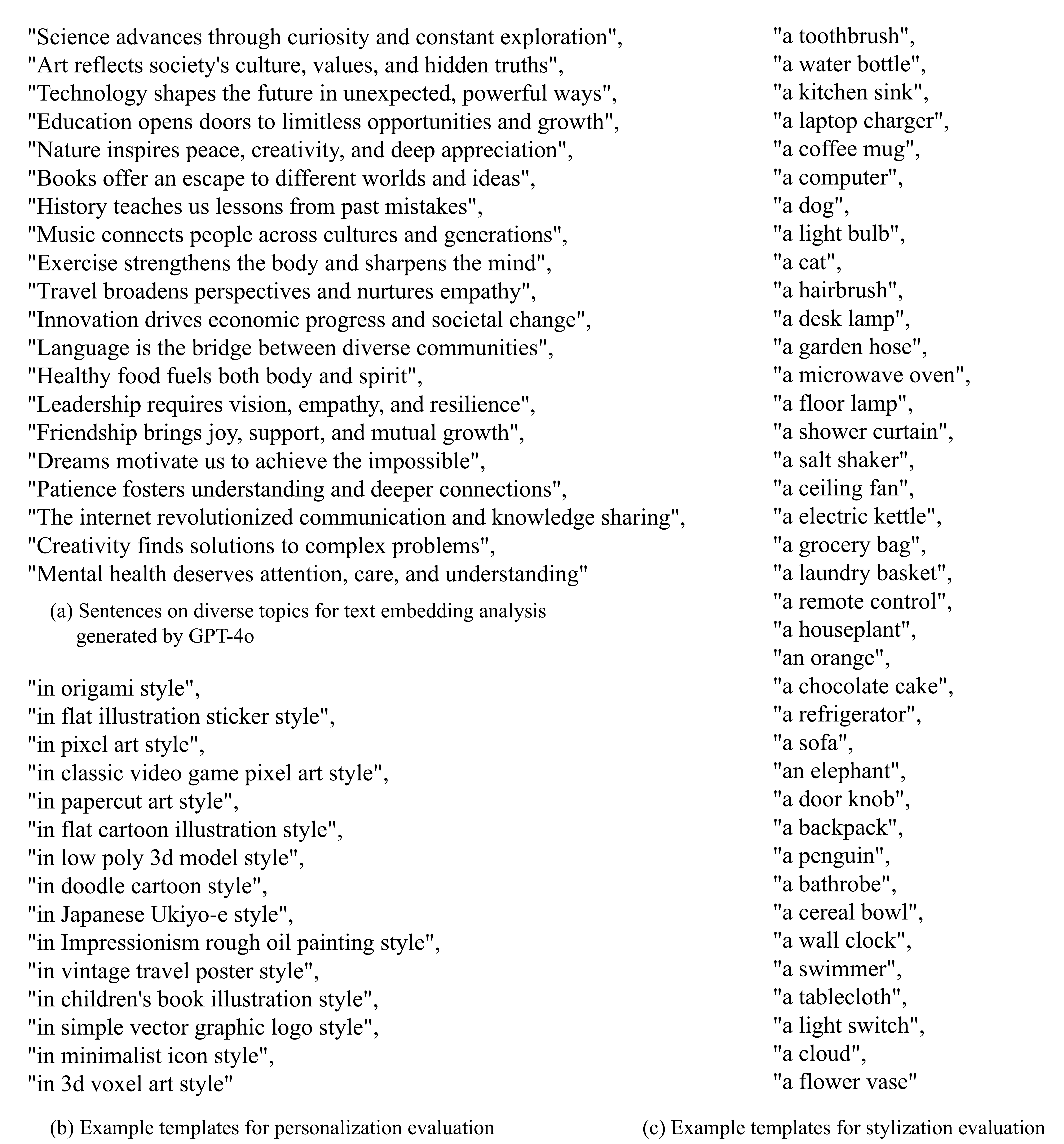}
    \caption{Templates used for (a) embedding analysis and (b), (c) performance evaluation.}
    \label{fig:template}
\end{figure*}

We demonstrate that T2I generation results can be adjusted through text embedding projection and propose a framework applicable to various tasks using LoRA, such as personalization, stylization, and content-style mixing. Despite effectively preventing overfitting, challenges remain. \cref{fig:limitation} compares stylization results from DreamBooth and DECOR. In some cases, better results might involve incorporating detailed elements from the reference image (\eg, the circular background and plant decorations) into the generated image. In other words, evaluating style is subjective, and the importance of certain visual features varies depending on the user~\cite{csd}. While our method allows implicit control of such detailed visual elements by adjusting $\alpha$, as shown in \cref{fig:fig13_ablation} and \cref{fig:additional_ablation}, it has limitations in providing explicit control. Future research could focus on enabling precise control over individual elements within generated images.

\section{Extended related work}

\noindent \textbf{Semantic refinement through embedding projection in text feature space.} In the context of text classification, Qin \etal~\cite{projection_cls} proposed a method that projects feature vectors in a direction that is orthogonal to common feature vectors. This allows the model to distinguish class-specific features from non-discriminative ones, enabling it to effectively capture the essential features for classification. Grand \etal~\cite{projection_human} introduced the semantic projection technique to extract contextual features of objects from word embeddings. By defining semantic axes as the vector differences between antonyms, they orthogonally project word vectors onto these axes to determine the relative positions of the objects. These studies suggest that projecting text embeddings onto orthogonal spaces can facilitate semantic transformations.

\noindent \textbf{Modifying text embedding for image manipulation.} Active research has focused on the properties of text embeddings for visual manipulation tasks, including image editing or attribute binding. Li \etal~\cite{getwhatyouwant} proposed an image editing technique by applying SVD and using a softmax operation on the singular values to modify text embeddings. This approach uses the modified embeddings to adjust the attention map for editing the image. Zhuang \etal~\cite{magnet} defined positive and negative binding vectors based on the similarity of the initial and final padding token embeddings for the attribute binding. Both studies offer insights that modifying the input text condition in T2I models can guide the model to follow the correct synthesis path during image generation. In addition to these studies, we propose a method for projecting embeddings onto a space orthogonal to unwanted targets. To the best of our knowledge, the proposed DECOR is the first approach to demonstrate that refining text embeddings at the inference, without additional training, can improve performance in LoRA-based customization tasks.

\clearpage

\clearpage
\clearpage

\clearpage


\end{document}